\documentclass[sigconf]{acmart}

\usepackage[utf8]{inputenc}
\usepackage{bm}

\usepackage{bbding}
\usepackage{wasysym}
\usepackage{algorithm}
\usepackage[noend]{algpseudocode}
\usepackage{multirow}
\usepackage{mathrsfs}

\usepackage{stfloats}

\AtBeginDocument{%
  \providecommand\BibTeX{{%
    \normalfont B\kern-0.5em{\scshape i\kern-0.25em b}\kern-0.8em\TeX}}}

\copyrightyear{2022}
\acmYear{2022}
\setcopyright{acmcopyright}\acmConference[ASIA CCS '22]{Proceedings of the 2022 ACM Asia Conference on Computer and Communications Security}{May 30-June 3, 2022}{Nagasaki, Japan}
\acmBooktitle{Proceedings of the 2022 ACM Asia Conference on Computer and Communications Security (ASIA CCS '22), May 30-June 3, 2022, Nagasaki, Japan}
\acmPrice{15.00}
\acmDOI{10.1145/3488932.3497753}
\acmISBN{978-1-4503-9140-5/22/05}

\begin{document}

\fancyhead{}
\title{Model Extraction Attacks on Graph Neural Networks:
\\
Taxonomy and Realisation}

\author{Bang Wu}
\email{bang.wu@monash.edu}
\affiliation{%
  \institution{Monash University}
  \city{Melbourne}
  \country{Australia}
}
\author{Xiangwen Yang}
\email{wayne.yang@monash.edu}
\affiliation{%
  \institution{Monash University}
  \city{Melbourne}
  \country{Australia}
}
\author{Shirui Pan}
\email{shirui.pan@monash.edu}
\affiliation{%
  \institution{Monash University}
  \city{Melbourne}
  \country{Australia}
}
\author{Xingliang Yuan}
\email{xingliang.yuan@monash.edu}
\affiliation{%
  \institution{Monash University}
  \city{Melbourne}
  \country{Australia}
}


\begin{abstract}

Machine learning models are shown to face a severe threat from Model Extraction Attacks, where a well-trained private model owned by a service provider can be stolen by an attacker pretending as a client.
Unfortunately, prior works focus on the models trained over the Euclidean space, e.g., images and texts, while how to extract a GNN model that contains a graph structure and node features is yet to be explored.
In this paper, for the first time, we comprehensively investigate and develop model extraction attacks against GNN models. %
%
We first systematically formalise the threat modelling in the context of GNN model extraction and classify the adversarial threats into seven categories by considering different background knowledge of the attacker, e.g., attributes and/or neighbour connections of the nodes obtained by the attacker. 
Then we present detailed methods which utilise the accessible knowledge in each threat to implement the attacks. 
By evaluating over three real-world datasets, our attacks are shown to extract duplicated models effectively, i.e., $84\%$ - $89\%$ of the inputs in the target domain have the same output predictions as the victim model. 
\end{abstract}



\maketitle

\section{Introduction}
\footnotetext[1]{The code of the paper is released at https://github.com/TrustworthyGNN/MEA-GNN}

%
%
%
Graph data are ubiquitously used in many applications, e.g., social media, document collections, and rating networks~\cite{KlicperaBG19,jagielski20}. 
To substantially analyse the graphs, graph neural networks (GNNs), as graph-based machine learning (ML) models, have been increasingly explored and offered state-of-the-art performance~\cite{KlicperaBG19, wan2021contrastive, zhuneurips2020, GilmerSRVD17}. 
As known, a well-trained machine model is costly during the data gathering, training period, and is often considered as the intellectual property of its owner~\cite{zitnik18}. 
To cater for the demands, cloud/AI platforms, e.g., Amazon SageMaker and Google Cloud AutoML, provide privatisation deployments for model owners to sell their models with a licensing fee~\cite{simon_2019}. 
Besides, GNN models used for e-commerce recommendation~\cite{NiuLLXSDC20} also provide public API to vast customers. 
On the other hand, such commercialisation draws much attention to the security of the models.

\begin{figure}[t]
    \centering
    \includegraphics[width=0.44\textwidth, height=0.22\textwidth]{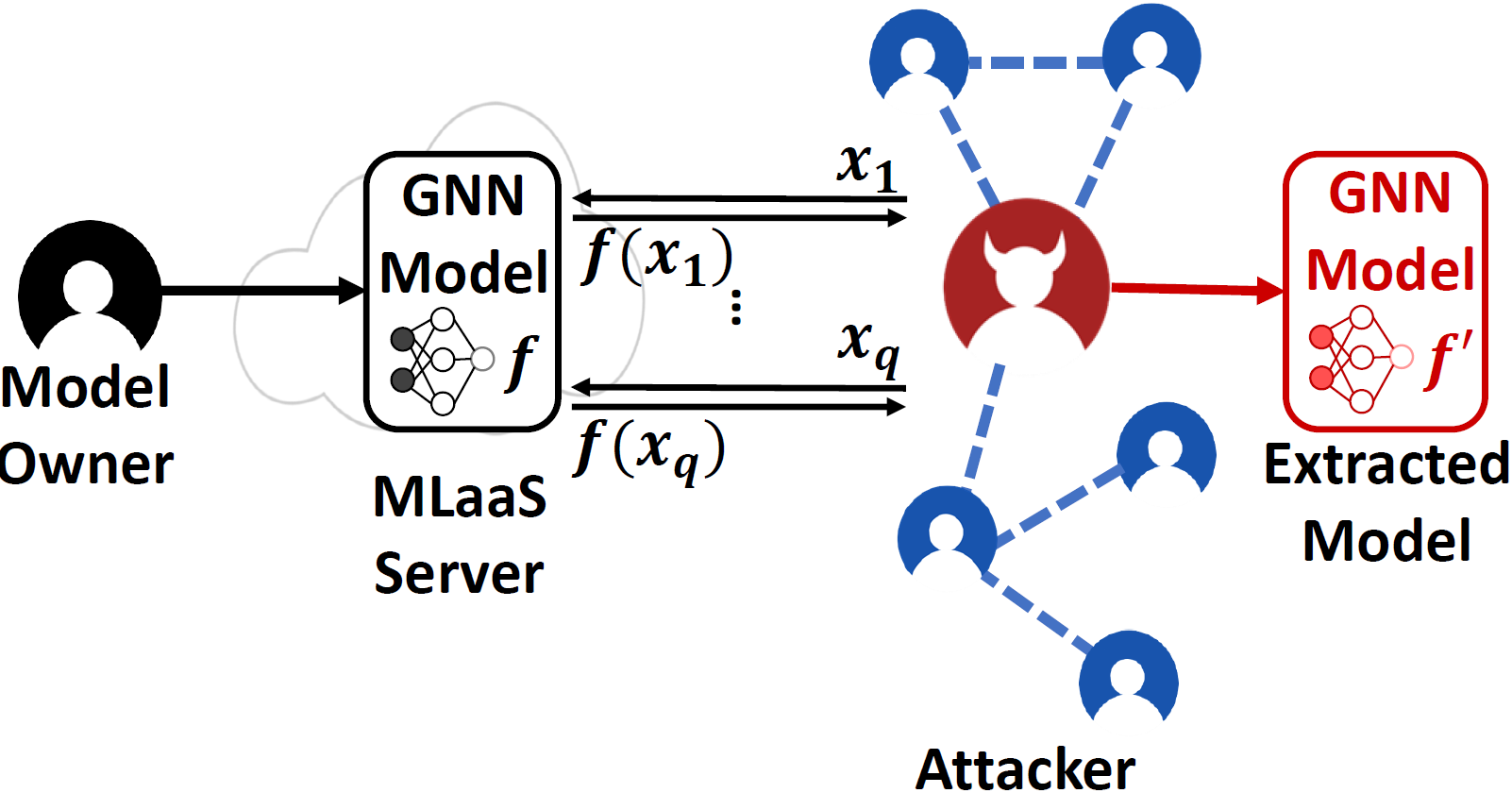}
    \caption{GNNs Model Extraction Attacks. A model owner provides a GNN model $f$ and the service of prediction queries. An attacker extracts a surrogate model $f^{'} \approx f$ based on the answers from the server.}
    \label{fig:overview}
\end{figure}
%

It has been demonstrated that attackers can steal ML models by Model Extraction Attack~\cite{Florian16,OhSF19}. 
Different from the adversarial attacks which aim at deducting the performance of ML systems, they propose to reconstruct a substitute model from the responses of generated queries to the target model. 
%
Because the input-output mappings of these queries contain sufficient information of the model prediction tasks, the extracted model can be quite similar to the target model, i.e., achieving comparable accuracy or generating the same prediction results as the target model. 
However, existing attacks only target the models with non-graph structures, e.g., MLP and CNN, while few studies focus on graph data.
The threats of GNN model extraction are still unclear.  
In this paper, we are the first to systematically explore and develop the model extraction attacks against GNNs. 
Specifically, referring to a private GNN model as the target, an extraction attacker attempts to construct a duplicated model with similar functionality via a sequence of queries from the nodes they obtained (aka \textit{attack nodes}).
Figure~\ref{fig:overview} illustrates an example of how our attack steals a GNN model on a node classification task of a social network. 
We consider an extraction attacker blending in with normal users in the network.
The attacker generates queries to the target model deployed in a machine learning as a service (MLaaS) and obtains the responses via its APIs.
%
The training of the duplicated model then utilises the information extracted from the input-output pairs. 


Unlike attacks to other neural networks, the extraction of GNNs requires knowledge in addition to the input-output mappings. 
An attacker targeting these graph-based classifiers needs to further consider the contributions of graph structures to classification tasks.
Different from the models for non-graph structured data, a GNN model predicts labels based on not only the input nodes but also their connections.
For example, the type of an image can be inferred by a CNN model individually, but the prediction of a node label will consider all the attributes of the node and other connected nodes.
Accordingly, such knowledge about the graph structure should also be taken into consideration during the model extraction.
%

The above specific requirement poses new challenges in developing model extraction attacks on GNNs. 
Existing attack strategies cannot be directly applied to GNNs since they leave out of graph structure during extraction. 
As mentioned, the attacker in GNNs should additionally gather the graph structure, but such knowledge sometimes cannot be obtained in real-world applications. 
%
For example, considering an online social network that contains both public and hidden user information, the attacker may have access to the public data, such as friendships (connections between users) but not to the private data, such as personal interests (attributes of users)~\cite{GuptaGJG13,GongL18,DuLZZXZS18}.
%
Therefore, how to design model extraction attacks with only missing or incomplete knowledge of the target model training graph is non-trivial.

To address the above challenges, we first propose a comprehensive framework for model extracting attacks against GNNs to understand and capture the capabilities of real-world attackers. 
%
We formalise the threat modelling by considering the attackers with diverse background knowledge in practice. 
The knowledge includes three dimensions: the attributes of the attack nodes, the partial graph consisting of the attack nodes, and an auxiliary sub-graph (shadow graph), including its graph structure and attribute information. 
This sub-graph can be exclusive to the graphs that are used to train the target models but having similar attributes and graph structure. 
Then, we characterise our attacks into seven different types of model extraction with or without this knowledge, and realise them using adaptive strategies. 
Specifically, if an attacker knows the graph structure but lacking node attributes knowledge, we design attribute synthesis algorithms to enrich the node attribute set and improve the attacks.
If the obtained graph structure is fragmented, we utilise the known attributes to construct a surrogate graph by graph structure generation methods.
%
%

The main contributions of our work are summarised as follows:
\begin{itemize}
    \item 
    To our best knowledge, for the first time, we systematically develop a series of GNN model extraction attacks that can steal a GNN model. 
    The extracted model behaves similarly to the target victim model.
    
    
    \item We propose a framework of threat modelling in the context of GNNs, which formalises and characterises the attacker's knowledge from three dimensions: node attributes, graph structure, and shadow sub-graph. 
    
    \item We define seven types of model extraction attacks under the above framework and realise them via adaptive attack strategies. 
    We implement each attack by utilising known background knowledge and constructing a surrogate training graph to build a duplicated model. 
    
    \item We evaluate our attacks over three real-world datasets. 
    The experiments confirm that, our attacks can effectively extract a duplicated model that is similar to the target model.
    Most of the duplicated models achieve nearly equal accuracy as the target, and more than $85\%$ prediction results from them are the same as the target.  
    
\end{itemize}

The rest of the paper is organized as follows. 
Section~\ref{sec:related_works} discusses related studies about the model extracted attacks targeting at ordinary Deep Learning system's privacy, and other adversarial attacks against GNNs. 
Section~\ref{sec:prel} introduce the preliminaries of our design.
Section~\ref{sec:attack_set} propose the goal, knowledge and the taxonomy of the attacks.
Section~\ref{sec:attack_method} introduces the detailed attack methodologies for each of our attacks.
Section~\ref{sec:exp} shows our experimental results.
Section~\ref{sec:conclusion} provides the summary and conclusion of our paper. 


\section{Related Work}
\label{sec:related_works}
\noindent \textbf{Model Extraction Attacks. }
Model extraction attacks targeting the confidentiality of ML systems have become paramount and have been explored in lots of studies~\cite{Smitha19,Soham19,Tribhuvanesh19,GaoLWCZ21,GongWCYJ20,Reith0T19}. 
Tram{\`{e}}r \textit{et al.}~\cite{Florian16} propose the first model extraction attacks against the linear ML models via Prediction APIs.
They reconstruct the model by solving the equations built by the queries, and the labels or confidence values.
They also use a path finding approach to attack the decision tree models.
Later, more studies consider attacking complex ML models, e.g., Neural Networks.
Milli \textit{et al.}~\cite{Smitha19} provide a gradient-based algorithm that extracts a two-layer ReLU network by carefully choosing the query inputs.
Pal \textit{et al.}~\cite{Soham19} demonstrate an attack on DNNs for both image and text classification tasks with active learning strategies.
Orekondy \textit{et al.}~\cite{Tribhuvanesh19} propose the attack by training a "knockoff" model which aims to match or even exceed the accuracy of the target model by generating query-prediction pairs.

Several approaches have also been proposed to defend against model extraction attacks, but they are not suitable for our attacks.
Some of them propose to hide or add noise to the output probabilities while maintaining the label outputs~\cite{Florian16,Chandrasekaran18,Taesung19}. 
But they are less effective in facing label-based extraction attacks like our design.
Others try to monitor each query and differentiate the adversarial ones by analysing the input distribution or the output entropy~\cite{Juuti19,Kesarwani18}. 
However, they do not consider the graph structure and are not optimised for GNN models.

\noindent \textbf{Attacks on Graph Neural Networks. }
Many studies have explored the vulnerability in GNNs. 
Most of them are adversarial attacks that target the integrity of the GNN systems~\cite{Zugner19,Wu0TDLZ19,XuXP21,abs-2007-08273,WangG19}. 
Z{\"{u}}gner \textit{et al.}~\cite{Zugner19} propose a scalable greedy approximation scheme to find the perturbation attacking the node classification GNNs.
They evaluate both node attribute and graph structure perturbation and compare their effectiveness.
Zhang \textit{et al.}~\cite{ZhangWY0WYP21} present the a transferable attacks against the graph-level GNNs. 
Wang \textit{et al.}~\cite{WangG19} generate the adversarial inputs by adding fake nodes into existing graphs without manipulating the existing connections. 
Zhang \textit{et al.}~\cite{Zhang19} propose a collection of data poisoning attack strategies, by manipulating the facts on the target graph. 
Xu \textit{et al.}~\cite{XuXP21} and Zhang \textit{et al.}~\cite{ZhangJWG21} propose the backdoor attacks by poisoning the training graph. 
Li \textit{et al.}~\cite{Li20} study the attack on the graph learning-based community detection models via hiding a set of nodes based on their surrogate model. 

Recent studies also draw attention to attacking the confidentiality of GNNs. 
A set of advanced attacks called membership inference attacks aim to infer whether a data sample has been used during the target model training~\cite{VidyalakshmiWC16,GongL18,JiaG18}.
Besides, He \textit{et al.}~\cite{abs-2005-02131} apply link stealing attacks against GNNs which can infer whether there is a link between two nodes on their training graph. 
Wu \textit{et al.}~\cite{wypy2021miagnn} propose the membership inference attacks against the graph-level GNN classifiers.
Most of them target the components of the graph rather than the GNN models, and our work aims to fill this gap. 

\section{Preliminaries}
\label{sec:prel}
\begin{table}[t]
    \centering
    \begin{tabular}{ll}
    \toprule
      Notation  & Explanation \\
    \toprule
    $G$ & An attributed graph of target model training \\
    $V$ & Node set of $G$ \\
    $E$ & Edge set of $G$ \\
    $X$ & Attribute set of $V$ \\
    $Y$ & Label set of the nodes $V$ \\
    $f_{\theta}(.)$ & A node classification model with parameters $\theta$ \\
    $P$ & Prediction result set of  $f_{\theta}(v_i)$ for every node $v_{i} \in V$ \\
    \midrule
    $G'$ & A shadow graph with the same domain as $G$ \\
    $V_{\mathcal{A}}$ & Attack node set \\
    $V_{\mathcal{A},k-hop}$ & $k$-hop neighbour node set of the attack nodes $V_{\mathcal{A}}$ \\
    $E_{\mathcal{A}}$ & Connections among the attack nodes $V_{\mathcal{A}}$ \\ 
    $E^*_{\mathcal{A}}$ & Synthetic connections among the attack nodes $V_{\mathcal{A}}$ \\
    $E_{\mathcal{A},k-hop}$ & $k$-hop neighbour connections of the attack nodes $V_{\mathcal{A}}$ \\
    $X_{\mathcal{A}}$ & Attributes of the attack nodes $V_{\mathcal{A}}$ \\
    $X^*_{\mathcal{A},k-hop}$ & Synthetic attributes of the $k$-hop neighbours \\
    $D_i$ & Degree of the node $v_i$ \\
    \bottomrule
    \end{tabular}
    \caption{Notations}
    \label{tab:notation}
\end{table}

In this section, we review a typical task of GNNs, and then proceed with the architecture of the target models, which is prevalently used to evaluate the attacks in GNNs. 
%
Our attacks under these typical scenarios can also be extended to the GNN models with other architectures. 

\noindent \textbf{Node Classification. } Given an attributed graph $G=(V,E,X)$, a set of nodes $V$ with node features $X$ are connected by a set of edges $E$.
A node classification model $f(.)$ can assign node labels $Y$ to each node in $V$ corresponding to both their node features and the graph structure.
We denote a classifier with parameters $\theta$ as $f_{\theta}(.)$.
The classification result for this model on a node $v_i$ (where $v_i \in V$) in $G$ is designated as $p_i=f_{\theta}(v_i)$.
For a well-trained GNN model, $p_i$ is expected to be the same as $y_i$, the corresponding label of $v_i$.

\noindent \textbf{Graph Convolution Networks. } In this paper, we consider a general graph convolution network (GCN) \cite{KipfW17}, indicated by $f_{\theta}(.)$, for the node classification task. 
GCN contains convolution layers that aggregate attribute information from the neighbour nodes. 
The equation for a two-layer GCN is defined as:
\begin{equation}
    f_{\theta}(A,X)=softmax(\hat{A} \cdot ReLU(\hat{A} \cdot X \cdot W^{(0)}) \cdot W^{(1)}), 
    \label{eqt:GCN}
\end{equation}
where $\hat{A} = \hat{D}^{-1/2} \widetilde{A} \hat{D}^{-1/2}$ denotes the normalised adjacency matrix, $\widetilde{A} = A + I_{N}$ denotes adding the identity matrix $I_{N}$ to the adjacent matrix $A$. $\hat{D}$ is the diagonal matrix with on-diagonal element as $\hat{D}_{ii} = \sum_{j} \widetilde{A}_{ij}$. $W^{(0)}$ and $W^{(1)}$ are the weights of first and second layer of GCN, respectively. $ReLU(0, a) = max(0, a)$ is adopted.
The notations we use throughout the paper are summarised in Table~\ref{tab:notation}.
Considering the model extraction attacks against GCNs, the parameters of the targeted model are $W$=\{$W^{(0)}, W^{(1)}\}$. 
As a result, the goal of the attacks targeting a GCN model becomes reconstructing the weights $W$.


\section{Attack Statement}
\label{sec:attack_set}
\subsection{Adversary's Goal}
Considering an MLaaS system, a private model provided by an entity can be deployed to the cloud server.
This server provides a query interface to the clients, while the clients can issue queries to the server and receive the responses. 
The model extraction attack aims to utilise the information derived from these input-output query pairs, extract the knowledge about the private model, and reconstruct a surrogate model. 

Formally, we consider a GNN model $f_{\theta}(.)$ trained on an attributed graph $G = (V, E, X)$ for a node classification task. 
The model extraction attack attempts to reconstruct a surrogate model $f_{{\theta}^{'}}(.)$ such that ${\forall}v_i{\in}V, f'_{{\theta}^{'}}(v_i)\approx f_{\theta}(v_i)$, where $V$ is a set of all the nodes in the graph and $v_i$ is one of the nodes.

We define a successful attack as one in which the attacker constructs a model that achieves similar performance to the target model (e.g. achieving similar accuracy at the testing set, or providing similar output predictions). 
To achieve this objective, the extracted model parameters do not need to be identical to the targeted ones. 
Namely, the model weights, or even the structure of the model, may be different than the target models once they have the same performance as the target models. 
The above definition is consistent with the one used in the ordinary DNN system. 
In practice, it is sufficient to harm the privacy of GNNs if a model with similar performance is extracted. 

\subsection{Adversarial Knowledge}
\label{sec:knowledge}

\begin{table}[]
    \centering
    \begin{tabular}{c|ccc|c|ccc}
    \toprule
        Attack & $X$ & $A$ & $G'$ & Attack & $X$ & $A$ & $G'$ \\
    \toprule
      —— & \Circle & \Circle & \Circle & Attack-3 & \Circle & \Circle & \CIRCLE \\
    \midrule
      Attack-0 & \LEFTcircle & \LEFTcircle & \Circle & Attack-4 & \LEFTcircle & \LEFTcircle & \CIRCLE \\
    \midrule
        Attack-1 & \LEFTcircle & \Circle & \Circle & Attack-5 & \LEFTcircle & \Circle & \CIRCLE \\
    \midrule
        Attack-2 & \Circle & \CIRCLE & \Circle & Attack-6 & \Circle & \CIRCLE & \CIRCLE \\ 
    \bottomrule
    \end{tabular}
    \caption{Taxonomy of the proposed threat model. $X$ represents the target dataset’s nodes attributes, $A$ represents the target dataset’s graph structure, $G'$ represents a shadow graph, and \CIRCLE/\LEFTcircle/\Circle means the attacker has complete/partial/no  knowledge.}
    \label{tab:attack_taxonomy}
\end{table}

Attackers with diverse background knowledge can apply the model extraction attacks at different levels.
In this paper, we tackle the most challenging adversarial setting: black-box attacks. Following the black-box assumptions in GNN attacks~\cite{ChangRXHZC0H20,0001DM20}, the attacker may obtain the adjacency matrix, attribute matrix, and output of the victim model, while the model parameters, labels and output probability are unknown.
In practice, it is reasonable that the attacker may only get access to a set of attack nodes, i.e., a subset of the nodes in the entire graph~\cite{0001DM20}. 
Namely, they do not have full knowledge of the inputs and outputs. 
In this paper, we propose different attack methods considering various adversarial background knowledge. 
They are characterised by three dimensions as below:

    
    \noindent \textbf{Nodes' Attributes $X$ of the Target Training Graph.}
    This characterises how much the attacker knows about the attributes $X$ of the nodes $V$ in the graph $G$ used to train the target model. 
    Generally, the attacker can have full access to the attack nodes they obtain, and can directly collect the node attributes for applications that store them in each node, e.g., users' profiles in a social network service are accessible by end users. 
    On the contrary, node attributes may not be obtained in some classification tasks.
    Note that, the attributes of other nodes which are not compromised should be invisible to the attacker consistently~\cite{WangJCYZ16,Wu0TDLZ19}.
    Therefore, we consider that the attacker cannot obtain the attribute knowledge except the attack nodes.
    
    \noindent \textbf{Graph Structure $A$ of the Target Training Graph.} 
    This characterises how much the attacker knows the graph structure of the target graph $G$. 
    Unlike the attributes which contain only the information from one node, the graph structure presents the relationship among multiple nodes.
    An attacker knowing the edges of the attack nodes can construct a sub-graph consisted of both the attack nodes and their neighbours. 
    Besides, while the node attributes are considered as private data, the connections such as friendships and following relations can be public.
    The attacker can reconstruct the target graph by crawling such public information~\cite{ChauPWF07,CataneseMFFP11}. 
    He can also utilise graph structure reconstruction methods~\cite{abs-2005-02131,abs-2010-00906} to obtain this knowledge.
    
    \noindent \textbf{Shadow Dataset $G'=(V',E',X')$.} 
    This represents a dataset in the same domain as the target dataset. 
    An example could be a scenario when the target dataset and shadow dataset are from the same large network but different sub-graphs or communities~\cite{GaoWJ18,ChiangLSLBH19}.
    In practice, the model owner may only have the privilege or the capability to train their model based on the sub-graph in an extensive network. 
    We assume the attacker may also have this privilege for another sub-graph as prior attack settings ~\cite{abs-2005-02131,GongL16,ShokriSSS17}.

\subsection{Attack Taxonomy}

\begin{figure*}
    \centering
    \includegraphics[width=1\textwidth, height=0.37\textwidth]{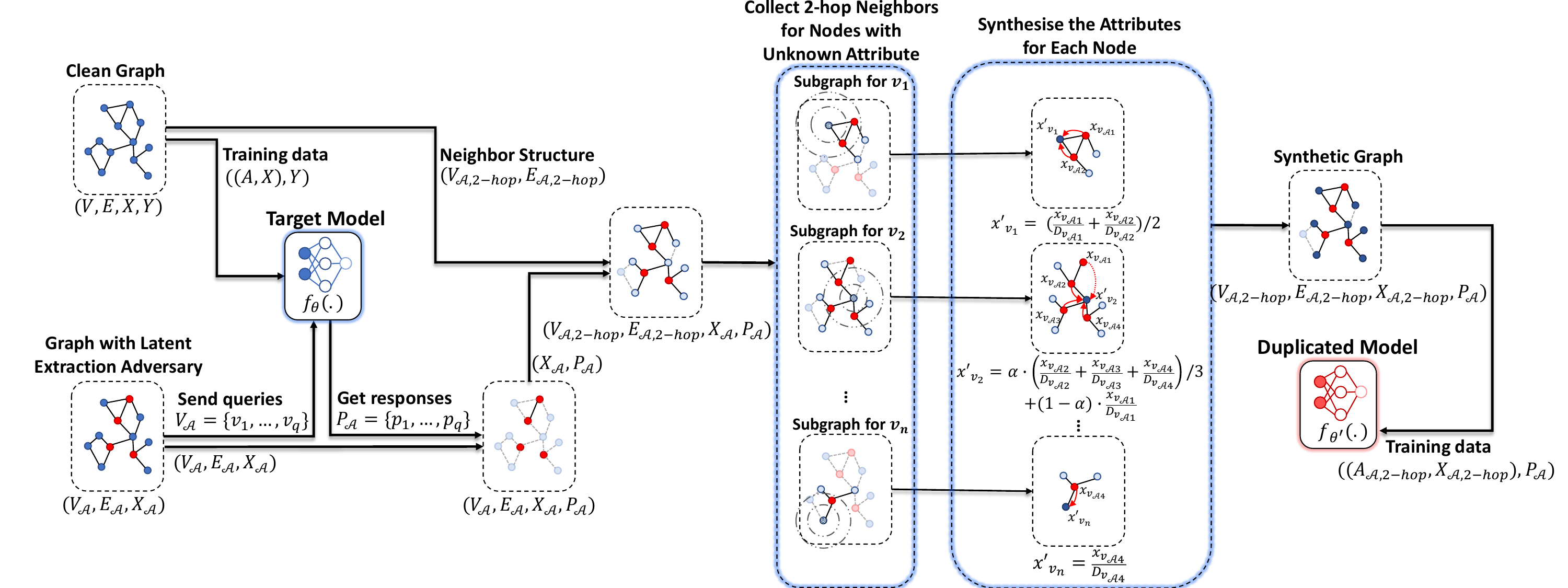}
    \caption{Illustration of Attack-0. 
    After obtaining the query responses $P_{\mathcal{A}}$ from the target model $f_{\theta}$, and the neighbour connections $(V_{\mathcal{A},2-hop},E_{\mathcal{A},2-hop})$ from the target graph, the attacker synthesises the attributes $X'_{\mathcal{A},2-hop}$ for the neighbour nodes of the attack nodes.
    The combination of the attack nodes with known attributes and labels, and the synthetic nodes with generated attributes
    $(V_{\mathcal{A},2-hop},E_{\mathcal{A},2-hop},X'_{\mathcal{A},2-hop},P_{\mathcal{A}})$ are used to train the duplicated GNN model $f_{\theta^{'}}$ via semi-supervised learning. }
    \label{fig:overview_attack_0}
\end{figure*}

Combining the above three dimensions, the knowledge of the attackers can be denoted as $(X, A, G')$. 
Based on whether the attacker has or has no knowledge of each item, we categories seven attacks by considering among the total eight cases. 
Note that we do not consider the case where the attacker has knowledge of neither three dimensions. 
The settings and scenarios for each of them are summarised as Table~\ref{tab:attack_taxonomy}. 

\noindent \textbf{Attack-0$(X, A, *)$: }
We present our first attack when the attacker has access to the attributes, connections, but no a shadow dataset of the target GNNs. 
Here, we assume a more practical setting, where the attacker has only partial knowledge of the node attributes and their connections. 
Specifically, only the attributes and neighbour sub-graph structure of the attack nodes are known by the attacker. 
In practice, it considers the scenarios when the attacker can obtain some attack nodes among the target graph. 
For example, the attacker can create several accounts in a financial credit network so they can know their own profiles (attributes) and transactions to others (connections to neighbour). 
As for the setting where the attacker has full knowledge of all the node attributes or complete knowledge of the graph structure, is a stronger adversarial assumption than Attack-0. 

\noindent \textbf{Attack-1$(X, *, *)$: }
We now consider a more strict case when the attacker has access to the attributes, but neither access to the connections nor a shadow dataset of the target GNNs. 
In practice, this may represent the scenarios when the attacker can not fully control the attack nodes. 
For example, he can not create his own malicious accounts but obtain limited information (i.e. only the profile rather than the transactions) by exploiting the vulnerability of the financial credit systems. 
Namely, only the attributes of the attack nodes (partial knowledge of the node attributes) are obtained by the attacker. 

\noindent \textbf{Attack-2$(*, A, *)$: }
Similar to Attack-1, we consider another case when the attacker has access to the connections but no the attributes or the shadow graph. 
Such type of attack can be used in the GNN systems where the connections are public but the node attributes are private. 
For example, in a social network, the profiles for each users (node attributes) are private data which are protected, while the relationship among the users (connections) are often public which are easier to obtain. 
The attacker can manage to crawl the information from the public data and reconstruct the entire graph connections. 

\noindent \textbf{Attack-3$(*, *, G')$: }
We then consider a scenarios when the attacker only obtain a shadow graph data. 
As we introduced in Section~\ref{sec:knowledge}, the shadow graph can be a different sub-graph or community from the same entire graph as the target graph. 
In practice, this represent when the attacker have overall knowledge about the target training graph or only have limited privilege or capability on an extensive network the same as the target GNN developer. 

\noindent \textbf{Attack-4$(X, A, G')$: }
This type of attack consider the scenario which is the combination in Attack-3 and Attack-0. 
The attacker can have access to both the attributes and neighbour connections to the attack nodes in the target graph, as well as the shadow dataset. 
In practice, this represents when the attacker successfully obtain the attack nodes while having knowledge on a shadow graph. 

\noindent \textbf{Attack-5$(X, *, G')$: }
We then consider the scenario which is the combination in Attack-3 and Attack-1. 
The attacker can have access to only the attributes and the shadow dataset, but not the connections. 
Again we assume that the attacker can only obtain the attributes of the attack nodes. 

\noindent \textbf{Attack-6$(*, A, G')$: }
We finally consider the case which is the combination in Attack-3 and Attack-2. 
The attacker can have access to only the connections and the shadow dataset, but not the attributes. 

\noindent \textbf{Remark: }
Note that there is a cause where the attacker has no information about the attributes, connections, or task domain of the target dataset (i.e., the knowledge is represented as $(*,*,*)$). 
In this case, neither the input graph nor the model parameters are known to the attacker, while only the final prediction labels are exposed. 
As a result, the attacker is not able to gain any information regarding the inputs of the target GNNs, thus making it difficult to discover a link between inputs and outputs. 
In this paper, we do not focus on extraction attacks based on this assumption, and leave it as an open problem for future work. 

\section{Attack Realisation}
\label{sec:attack_method}



\subsection{Attack-0}

\begin{algorithm}[t]
\caption{Algorithm for Attack-0}\label{alg:attack_0}
\begin{flushleft}
\hspace*{\algorithmicindent}\textbf{Input:} \\
\hspace*{\algorithmicindent} $q$ attack nodes' attributes $X_{\mathcal{A}}$ $=$ $\{x_{v_1}$, $x_{v_2}$, ..., $x_{v_q}\}$, $q$ attack nodes' query results $P_{\mathcal{A}}=\{p_{v_1}, p_{v_2}, ... p_{v_q}\}$, 
graph structure of the $2$-hop neighbour nodes set of the attack nodes $(V_{\mathcal{A},2-hop},E_{\mathcal{A},2-hop})$, 
adjustment factor $\alpha$. \\
\hspace*{\algorithmicindent}\textbf{Output:} \\
\hspace*{\algorithmicindent} Extracted Model $f_{\theta^{'}}(.)$. \\
\end{flushleft}
\begin{algorithmic}[1]
\State Generate adjacency matrix $A_{\mathcal{A},2-hop}$ for $(V_{\mathcal{A},2-hop},E_{\mathcal{A},2-hop})$
\State Initialise a empty attribute set $X'_{\mathcal{A},2-hop}$
\For {$v_i \in V_{\mathcal{A},2-hop}$}
    \If{$v_i \in V_{\mathcal{A}}$}
        \State $\backslash\backslash$ Keep the attribute of attack nodes
        \State Collect $x_{v_i}$ from $X_{\mathcal{A}}$
        \State Add $x_{v_i}$ to $X'_{\mathcal{A},2-hop}$
        \State Label $v_i$ as $y'_{v_i}$ according to  $p_{v_i}$ in $P_{\mathcal{A}}$.
    \Else

        \State $\backslash\backslash$ Gather knowledge from the 1-hop neighbours
        \State Initialise a empty attribute set $X'_{v_i,1-hop}$
        \For {$v_j \in V_{v_i,1-hop}$}
            \If{$v_j$ is in $V_{\mathcal{A}}$}
                \State $x'_{v_j}=x_{v_j}/D_{v_j}$
                \State Add $x'_{v_j}/D_{v_j}$ to $X'_{v_i,1-hop}$
            \EndIf
        \EndFor
        \State $\backslash\backslash$ Gather knowledge from the 2-hop neighbours
        \State Initialise a empty attribute set $X'_{v_i,2-hop}$
        \For {$v_j \in V_{v_i,2-hop}$ and $\notin V_{v_i,1-hop}$}
            \If{$v_j$ is in $V_{\mathcal{A}}$}
                \State $x'_{v_j}=x_{v_j}/D_{v_j}$
                \State Add $x'_{v_j}/D_{v_j}$ to $X'_{v_i,2-hop}$
            \EndIf
        \EndFor
        \State $\backslash\backslash$ Synthesise attribute based on the 2-hop neighbours
        \State $x'_{v_i}=\alpha \cdot $mean$(X'_{v_i,1-hop})+ (1-\alpha) \cdot $mean$(X'_{v_i,2-hop})$
        \State Add $x'_{v_i}$ to $X'_{\mathcal{A},2-hop}$
    \EndIf
\EndFor
\State Train 2-layer GCN $f_{\theta^{'}}(.)$ based on $(X'_{\mathcal{A},2-hop},A_{\mathcal{A},2-hop},Y_{\mathcal{A}})$
%
\end{algorithmic}
\end{algorithm}

We first consider a scenario where the attacker obtains a set of attack nodes $V_{\mathcal{A}}$ and has both access to their attributes $X_{\mathcal{A}}$ and neighbour sub-graph structure $A_{\mathcal{A},k-hop}$. 
These attack nodes are randomly chosen among the total node set $V$ to imitate the real-world scenarios where every node in the victim graph can be a potential attack node.
%

To extract the target models, the attacker intends to generate a graph for the duplicated model training.
We call it \textit{an attack graph} in the rest of our paper.
The attack graph consists of the node attributes, graph structure, and node labels.
The attacker attempts to obtain or generate the above items based on their adversarial knowledge.
Specifically, the attack graph for the extracted model training can be built by three steps gathering each of the above items.
Figure~\ref{fig:overview_attack_0} shows a procedure for getting this attack graph.

\noindent \textbf{Issuing queries and obtaining labels. }
In our assumptions, the attackers can obtain the attribute and query results of the attack nodes.
The results of the response queries from the attack nodes can be considered as their node labels.
Hence, they are utilised as the labelled nodes with known attributes to train a duplicated model for the node classification task.

 \noindent \textbf{Gathering neighbour connections. }
Knowing input attributes, output predictions, and connections among the attack nodes, the attacker can naturally employ supervised learning to train the duplicated model.
However, the predictions of our attack nodes are also affected by their neighbours.
Training the model by the attack nodes isolated among the graph will desert the impacts from the neighbours of the attack nodes and reduce the attacker performance. 
Therefore, our design should rationally consider the attributes of the neighbours around these nodes.
Specifically, the attacker will gather the connections among the attack nodes and their neighbours, which are considered as the graph structure of the attack graph.

\noindent \textbf{Synthesising attributes for in-accessible nodes. }
In our assumptions, the attacker only knows the attributes and query results of the attack nodes.
Thus, the attacker needs to synthesise the attributes for the neighbour nodes with unknown attributes. 
In practice, most nodes have similar attributes as their neighbours~\cite{abs-2009-13504,GongL16,ChenLSL20}.
Based on this observation, the synthetic attributes can be the combination of their neighbour nodes' attributes. 
Formally, to synthesise the attributes of a target node $x'_{v_i}$, the attacker first gathers all its known-feature neighbours, including $n$ 1-hop nodes $\{v_{1,1-hop},...,v_{n,1-hop}\}$ and $m$ 2-hop nodes $\{v_{1,2-hop},...,v_{m,2-hop}\} \subset V_{\mathcal{A}}$.
For each of them, the impact to the targets can be represented as $v_{j,k-hop}/D_j$, where $D_j$ represents the degree of this neighbour node $v_j$. 
Considering an adjustment factor $\alpha$ to balance the effects from one or two hops nodes, the attacker synthesises the feature of the target node as:
\begin{equation}
    x'_{v_i} =  \alpha \sum_{j=1}^{n} \frac{x_{v_j,1-hop}}{nD_j} + (1 - \alpha) \sum_{j=1}^{m} \frac{x_{v_j,2-hop}}{mD_j}
\end{equation}

\noindent \textbf{Learning the extracted model. }
After generating the attributes for these nodes, the attacker can obtain a graph that includes all attack nodes and their neighbours with the known or synthetic attributes, and then train a node classification GNN model as the extracted model.
Note that the attacker does not label the synthetic nodes.
Unlike the labels of the attack nodes that come from the query responses, the synthetic nodes are inaccessible to the attacker.
He can neither modify their attributes nor send queries to the target models.
As a result, only the attack nodes can be labelled and the extracted models are trained via semi-supervised learning. 
The overall process of the Attack-0 is shown as Algorithm~\ref{alg:attack_0}.

\subsection{Attack-1}
\begin{figure}[t]
    \centering
    \includegraphics[width=0.45\textwidth, height=0.22\textwidth]{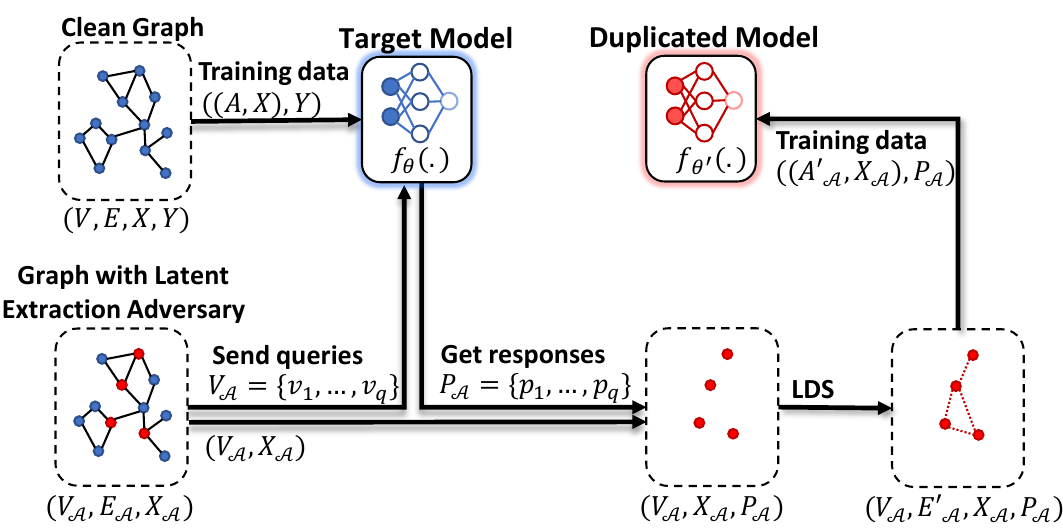}
    \caption{Illustration of Attack-1. 
    After obtaining the query responses $P_{\mathcal{A}}$ from the target model $f_{\theta}$, the attacker can only obtain discrete nodes with their attributes $X_{\mathcal{A}}$.
    A synthetic graph can be generated based on these attributes via graph generation method LDS~\cite{Franceschi19}.
    Then the attack nodes with attributes and labels and the synthetic graph structure $(V_{\mathcal{A}},E'_{\mathcal{A}},X_{\mathcal{A}},P_{\mathcal{A}})$ are used to train the duplicated GNN model $f_{\theta^{'}}$ via supervised learning. }
    \label{fig:overview_attack_1}
\end{figure}

We then intensify the restriction to the attacker and consider the case when the attacker has only knowledge about the attributes of the attack nodes $X_{\mathcal{A}}$. 
For this type of attack, the attacker also needs to first generate an attack graph for the extracted model training.
Compared with Attack-0, since the graph structure is unknown, the attacker needs to generate the connections between the nodes. 
A procedure of the attack is shown in Figure~\ref{fig:overview_attack_1} and the attack graph is generated as follows. 

\noindent \textbf{Issuing queries and obtaining labels. }
Similar to Attack-0, the attributes and the query responses of the attack nodes can be used as the labelled nodes in the attack graph.

\noindent \textbf{Synthesising connections among attack nodes. }
Different from Attack-0, the graph structure is unknown to the attacker. 
If all the attack nodes are deemed to be isolated, the impacts from the neighbours of the attack nodes cannot be taken into considerations.
To solve this problem, the attacker needs to construct a substitute graph based on the known attributes.

Generally, the attributes of nodes in a graph and the connections among them are tightly correlated~\cite{AcharyaZ20,abs-2009-00203}.
Thus, it is possible to infer or reconstruct the graph structure based on the node attributes.
Based on this intuition, several prior studies about graph synthesis and generation have been developed to generate graphs~\cite{abs-1802-08773,abs-1803-03324,Franceschi19}.
Among others, we use a graph generation method called Learning Discrete Structures (LDS)~\cite{Franceschi19}. It can generate the graphs by considering their performance on classification problems, which meets the tasks of our target models. 
Therefore, given the attributes of the attack nodes, the attacker can synthesise the connections among them and use the synthetic structure as the attack graph.

\noindent \textbf{Learning the extracted model. }
After the above steps, the attacker can obtain a substitute graph with attack node attributes, corresponding prediction labels, and a generated graph structure.
Then he can use supervised learning to train the duplicated models.
Note that, due to the approximation of the edges distributions, the density of the generated graph can be set close to the target. 
Hence, most of the nodes generated via this method are not isolated and the attacker does not need to synthesise their neighbours as Attack-0.

\subsection{Attack-2}
\begin{figure}[t]
    \centering
    \includegraphics[width=0.49\textwidth, height=0.23\textwidth]{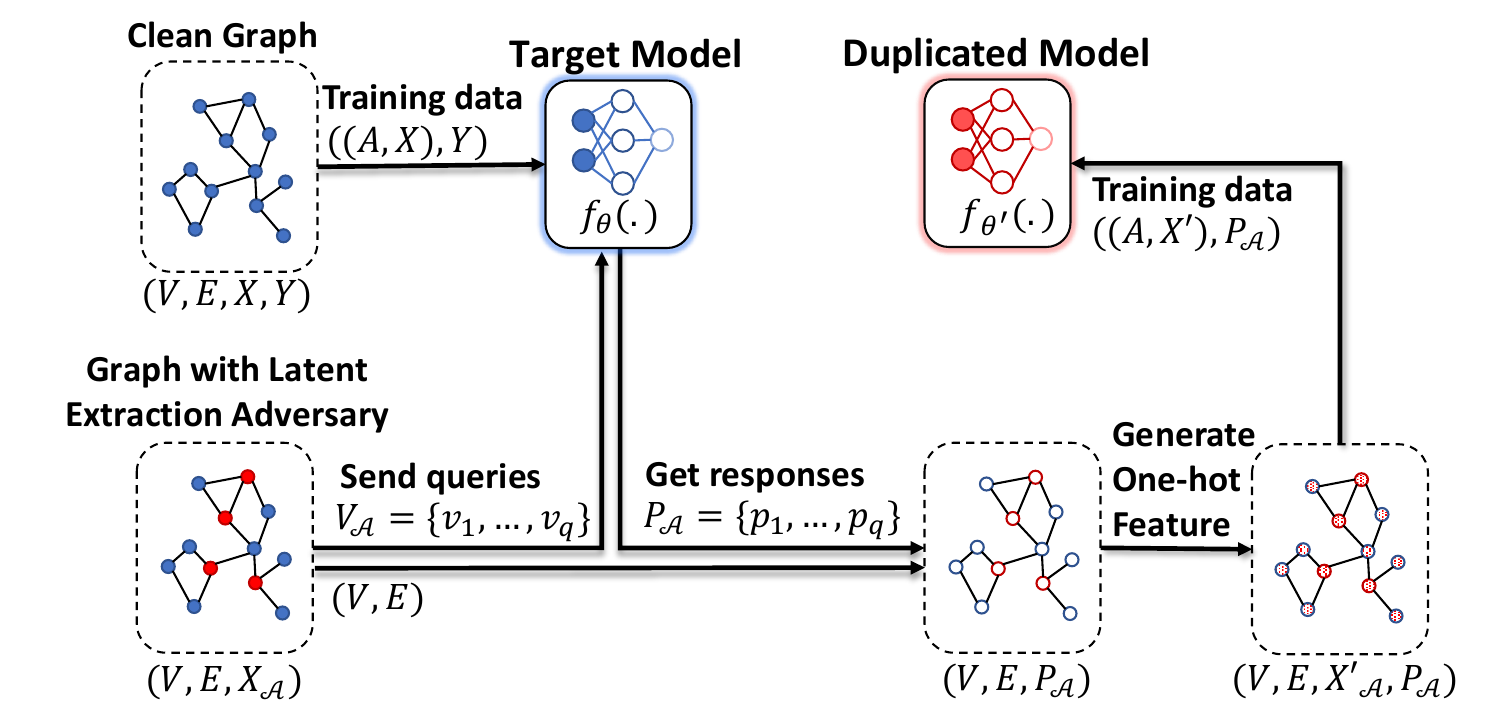}
    \caption{Illustration of Attack-2. 
    After obtaining the query responses $P_{\mathcal{A}}$ from the target model $f_{\theta}$, the attacker can only obtain the graph $(V,E,P_{\mathcal{A}})$ without node attributes.
    The attacker assigns one-hot vectors to every node as their synthetic attributes $X'$ based on their graph index.
    Then the attack nodes with attributes and labels and the synthetic graph structure $(V,E,X',P_{\mathcal{A}})$ are used to train the duplicated GNN model $f_{\theta^{'}}$ via semi-supervised learning. }
    \label{fig:overview_attack_2}
\end{figure}

We consider another scenario when the attacker obtains the entire graph structure knowledge $A$ while having no access to any nodes in $V$ even for the attack nodes $V_{\mathcal{A}}$.
Namely, he cannot obtain the node attributes $X$.
As a result, he needs to build the attack graph by synthesising the attributes, as depicted in  Figure~\ref{fig:overview_attack_2}.
The detailed steps are:

\noindent \textbf{Issuing queries and labelling the attack nodes. }
Even though the attacker has no access to the node attributes, he can still obtain the responses and use them to label the attack nodes. 
After generating attributes, these labelled nodes can be used during the extracted model training.

\noindent \textbf{Gathering the target graph as attack graph. }
To reconstruct the target model, the attacker naturally utilises the entire known graph structure to build the attack graph.

\noindent \textbf{Assigning one-hot vectors as node attributes. }
Without any knowledge about the attributes, the attacker proposes to first synthesise them and build a surrogate training graph.
As discussed in Section~\ref{sec:attack_set}, the attribute of a node is also related to the entire graph structure and its position in it.
To synthesise attributes associated with the structure knowledge, the attacker uses the index of the nodes to generate \textit{one-hot vectors} as their attributes.
For example, the attribute of $v_1$ will be $[1, 0, 0, ..., 0]$ while for $v_2$ is $[0, 1, 0, ..., 0]$. 
These attributes represent the identity of the nodes and contain information about the graph structure.

Note that, generating arbitrary features does not satisfy the model training with graph structure since they might bear no resemblance to their actual attributes. 
Meanwhile, it is also hard to reconstruct the original attributes of the target graph.
As mentioned, the node attributes and the graph structure are tightly related.
The objective of our attacks is to extract the mapping from node attributes to the node labels based on the graph structure.
Thus, inferring the original attributes of the graph based on the graph structure and the node label or even only the node position can be considered as the reversed function of our target models. 
It is difficult to first learn this reversed mapping and then extract the target models.

\noindent \textbf{Learning the extracted model. }
After synthesising the attributes, the attacker can  construct a surrogate model by these attributes, prediction labels of the attack nodes, and the entire graph structure via semi-supervised learning.
Different from previous attacks, the inputs of the surrogate models are the \textit{one-hot vectors}. 
As a result, the nodes will be inferred via their indexes in the graph. 
Since the entire target graph structure is utilised, the extracted models can be used to classify all the nodes in the target graph as the target model.

\subsection{Attack-3}
We now consider the case when the attacker does not know the node attributes and their connections. But we assume the attacker has access to a shadow graph $G'=(V',E',X')$ defined in Section~\ref{sec:attack_set}. 
Under this adversarial assumption, the attacker has no knowledge about the target graph.
Therefore, the extraction can only refer to the shadow graph.
As introduced in Section~\ref{sec:attack_set}, the shadow graph has the same domain as the target.
Therefore, it is possible to utilise the knowledge from a shadow graph, i.e., using it as the attack graph.

Specifically, the attacker first gathers both the node attributes $X'$ and the graph structure $A'$ of a shadow graph.
He can also obtain the corresponding labels $Y'$ for some nodes in the shadow graph.
Then, this shadow dataset $D'=(X',A',Y')$ can be used to train a surrogate model via semi-supervised learning.
Since the dimensions of the node attributes from the graph with the same domain are also the same, the parameters of the surrogate model (the weights $W$) have the same size.
Therefore, these weights can be used as the extracted model which achieves similar functionality as the target model if the target and shadow graphs are in the same domain.

\subsection{Attack-4}
In this attack, the attacker is assumed to have access to the attack nodes $V_{\mathcal{A}}$ as Attack-0. 
Besides, he can collect a shadow graph $G'$ as Attack-3.
%
With both the background knowledge as Attack-0 and Attack-3, the attacker proposes to combine them together. 
In particular, an associated attack graph is built by combining the attack graphs for these two attacks.

The attacker first generates an attack graph consisting of the attacker and synthetics nodes with the same strategy as Attack-0.
Then, the shadow graph is set to be the second attack graph as Attack-3.
Since the attack nodes in the first attack graph are often not connected to the second attack graph, the attacker will not synthesise the connections between them.
It can avoid the negative impacts among the attack nodes and the shadow graph.
Hence, the associated attack graph consists of these two isolated graph components.  
After that, the attacker can train the extracted model on this associated attack graph based on all the known node attributes, their graph structure, and the labelled nodes from both the shadow graph and the attack nodes in the target graph.  


\subsection{Attack-5}
This attack considers the case where the attacker has access to a shadow graph $G'$ and also the attributes of the attack nodes $X_{\mathcal{A}}$ in the target graph.
The adversarial knowledge is from both Attack-1 and Attack-3.
Similar as Attack-4, the attacker can combine them to utilise the background knowledge of $X_{\mathcal{A}}$ and $G'$ in this attack. 

To implement the attack, a graph generation method is used to construct a structural graph for the attack nodes based on their attributes as Attack-1. 
The graph structure and the corresponding nodes with attributes and query responses consist of the first attack graph. 
The attacker again uses the shadow graph as the second attack graph.
Due to the same reason as Attack-3, the attacker will not synthesise connections between two attack graphs.
Then, two attack graphs are associated to build the combined attack graph for the training of the extracted model.

\subsection{Attack-6}

\begin{algorithm}[t]
\caption{Algorithm for Attack-6}\label{alg:attack_6}
\begin{flushleft}
\hspace*{\algorithmicindent}\textbf{Input:} \\
\hspace*{\algorithmicindent}
Shadow graph $G'=(V',E',X')$, graph structure $(V,E)$, 
\hspace*{\algorithmicindent}\textbf{Output:} \\
\hspace*{\algorithmicindent} Extracted Model $f_{\theta^{'}}(.)$. \\
\end{flushleft}
\begin{algorithmic}[1]
\State $\backslash\backslash$ Extract GCN as Attack-2 on target graph
\State Generate one-hot vector attributes $X_{one-hot}$ from $(V,E)$.
\State Train GCN $f_{attack\_2}(.)$ based on $(V,E,X_{one-hot},Y)$.
\State $\backslash\backslash$ Extract GCN as Attack-2 on shadow graph
\State Generate one-hot vector attributes $X'_{one-hot}$ from $(V',E')$.
\State Train GCN $f'_{attack\_2}(.)$ based on $(V',E',X'_{one-hot},Y')$.
\State $\backslash\backslash$ Extract GCN as Attack-3 on shadow graph
\State Train GCN $f'_{attack\_3}(.)$ based on $(V',E',X',Y')$.
\State $\backslash\backslash$ Posterior feature generation
\For {$v'_i \in V'$}
    \State $x'_{attack\_2,v'_i} = f'_{attack\_2}(v'_i)$
    \State $x'_{attack\_3,v'_i} = ,f'_{attack\_3}(v'_i)$. 
\EndFor
\State Stack $X'_{attack\_2}$ and $X'_{attack\_3}$ as $X'_{attack\_6}$.
\State $\backslash\backslash$ Build ensemble models
\State Train DNN $f_{attack\_6}(.)$ based on $(V',E',X'_{attack\_6},Y')$.
\State $f_{\theta^{'}}(.)=f_{attack\_6}(f_{attack\_2}(.),f'_{attack\_3}(.))$
\end{algorithmic}
\end{algorithm}

This attack considers the assumption that the attacker has no access to the node attributes but the entire graph structure knowledge $A$ and a shadow graph $G'$. 
Compared with Attack-2, the attacker can gather extra knowledge from the shadow graph. 
However, it is hard to directly utilise this node attribute knowledge and implement a similar design as Attack-2.
In Attack-2, the attributes of the nodes are one-hot vectors corresponding to the node indexes.
Thus, the dimension of the synthetic attributes will be the same as the number of nodes  which does not match the original attributes. 

To combine Attack-2 and 3, the attacker can build the ensemble models. 
Two models utilising different background knowledge via the methods as Attack-2 or Attack-3 are trained separately. 
Even though their inputs are different, both their outputs are the posteriors of the label of nodes.
An attack model is trained to predict the final labels based on two posteriors.
Specifically, the inputs of the attack model are the stack of the outputs from the two extracted models while the outputs are the final prediction labels.
Since the attacker cannot obtain the posteriors from the target models of our attacks, the attack models are first generated in the shadow graph.
The detailed processes are:


\noindent \textbf{Extracting models on the shadow graph. }
We extract a model on the shadow graph via Attack-2 by only using its graph structure knowledge. 
The inputs of this model are one-hot vector attributes.
Then, we extract another model on the shadow graph via Attack-3 by using its entire graph data.

\noindent \textbf{Training an attack model. }
Since the attacker can obtain all knowledge about the node attributes, graph structure, he can feed them into the two models generated above and gather their output posteriors.
Then the posteriors and their corresponding labels are used to train a simple MLP model to predict the final labels.

\noindent \textbf{Extracting the model on the target graph. }
After building the attack model on the two emulated models in shadow graph, the attacker can generate the real extracted models.
To utilise the structure knowledge of the target graph, the model is extracted via Attack-2.

\noindent \textbf{Building the ensemble models. }
After training the two extracted models for Attack-2, Attack-3 and the attack model, the attacker can set up the ensemble model.
The output posteriors of the two extracted models are fed into the attack model to generate the final predictions.

\section{Experiments}
\label{sec:exp}
In this section, we present a comprehensive set of experiments to evaluate our attacks.
We first introduce the experiment setting and then present the detailed results for each attack.

\subsection{Experimental Setup}
\noindent \textbf{Datasets. }
Three public datasets are used to evaluate our proposed attacks, including Cora, Citeseer, and Pubmed~\cite{KipfW17}.
All of them are benchmark datasets that are widely used for the evaluation of node classification models.
These three datasets are citation networks whose nodes represent the publications and edges are their citations.
The detailed statistics of the datasets are shown in Table~\ref{tab:dataset}.
\begin{table}[t]
    \small
    \centering
    \begin{tabular}{c|ccc}
    \toprule
        Datasets & Node Number & Edge Number & Class Number\\
    \toprule
      Cora & 2708 & 5429 & 7\\
    \midrule
      Citeseer & 3327 & 4732 & 6\\
    \midrule
      Pubmed & 19717 & 44338 & 3\\
    \bottomrule
    \end{tabular}
    \caption{Dataset Statistics}
    \label{tab:dataset}
\end{table}

\noindent \textbf{Datasets Configuration. }
We configure the datasets for our different attacks.
For Attack-0, Attack-1, and Attack-2 which do not contain the knowledge about the shadow dataset, we use the entire graph data to train the target models.
For the Cora dataset, we split the network into 140 (about $5\%$ of the total nodes) labelled nodes as training part, 300 ($11\%$) labelled nodes as validation, and the rest of them unlabelled.
Among the unlabelled nodes, we choose 1000 ($37\%$) of them as the testing sets.
For the Citeseer, we set 120 ($4\%$) labelled nodes as training part, 500 ($15\%$) labelled nodes as validation, and 1000 ($30\%$) unlabelled nodes as the testing set.
And for the Pubmed dataset, 60 ($0.3\%$) labelled nodes are used as the training part, while 500 ($2.5\%$) labelled nodes as validation. 
We again choose 1000 ($5.1\%$) unlabelled nodes as the testing set.

For Attack-3, Attack-4, Attack-5, and Attack-6 using shadow dataset, we split the network into two parts: the graph for target model training and the graph assumed to be known by the attacker.
To generate the shadow dataset, we first split the entire network into several communities by Clauset-Newman-Moore greedy modularity maximisation~\cite{clauset04}, and then divide them into two datasets.
For the Cora network, we generate the training graph for the target models which consists of 1408 (about $50\%$) nodes. 
For the Citeseer, the training graph for the target models has 1320 (about $40\%$) nodes. 
And for the Pubmed dataset, we set the training dataset for the target models to be the graph with 1408 (about $50\%$) nodes. 
The rest of them which consist of 1300 nodes are used as the shadow dataset. 
We split both the target and the shadow networks into training and testing parts. 
Other configurations for the datasets share the same settings as the datasets for Attack-0, Attack-1, and Attack-2.  


\begin{table*}[!t]
    \centering
    \begin{tabular}{c|ccc|ccc}
    \toprule
        Metrics & \multicolumn{3}{c|}{Accuracy} & \multicolumn{3}{c}{Fidelity} \\
    \midrule
      Dataset & Cora & Citeseer & Pubmed  & Cora & Citeseer & Pubmed \\
    \midrule
      Target Model & $0.816$ & $0.713$ & $0.800$ & -- & -- & -- \\
      Simple DNN (baseline) & $0.577\pm0.004$ & $0.596\pm0.004$ & $0.727\pm0.006$ & $0.590\pm0.010$ & $0.632\pm0.005$ & $0.761\pm0.005$ \\
      Attack-0 & \bm{$0.799\pm0.009$} & $0.684\pm0.016$ & $0.736\pm0.004$ & \bm{$0.896\pm0.008$} & \bm{$0.848\pm0.019$} & \bm{$0.890\pm0.007$} \\
      Attack-1 & $0.798\pm0.006$ & \bm{$0.708\pm0.007$} & \bm{$0.751\pm0.003$} & $0.825\pm0.007$ & $0.754\pm0.005$ & $0.857\pm0.003$ \\
      Attack-2 & $0.762\pm0.012$ & $0.548\pm0.004$ & $0.652\pm0.036$ & $0.809\pm0.006$ & $0.602\pm0.003$ & $0.728\pm0.035$ \\
    \midrule
      Target Model & $0.816$ & $0.697$ & $0.806$ & -- & -- & -- \\
      Attack-3 & $0.809\pm0.007$ & $0.692\pm0.004$ & $0.799\pm0.001$ & $0.790\pm0.005$ & $0.714\pm0.002$ & $0.818\pm0.009$ \\
      Attack-4 & $0.801\pm0.009$ & \bm{$0.708\pm0.002$} & \bm{$0.800\pm0.008$} & $0.790\pm0.011$ & \bm{$0.736\pm0.008$} & \bm{$0.837\pm0.004$} \\
      Attack-5 & \bm{$0.832\pm0.004$} & $0.699\pm0.001$ & $0.799\pm0.002$ & \bm{$0.807\pm0.002$} & $0.727\pm0.002$ & $0.818\pm0.003$ \\
      Attack-6 & $0.800\pm0.017$ & $0.649\pm0.017$ & $0.737\pm0.092$ & $0.791\pm0.019$ & $0.731\pm0.019$ & $0.813\pm0.086$ \\
    \bottomrule
    \end{tabular}
    \caption{Model accuracy/fidelity for all attacks on three different datasets. Attack-0, Attack-1, and Attack-2 target at the model trained in the entire dataset. Attack-3, Attack-4, Attack-5, and Attack-6 target the model trained in a sub-graph split from the entire graph. Best results are highlighted in bold. }
    \label{tab:attack_overall}
\end{table*}

\noindent \textbf{Evaluation Metric. }
We evaluate our attacks from two aspects based on the two different definitions about the similar performance of extracting the models following the evaluation methods of prior extraction attacks in DNNs~\cite{,Tribhuvanesh19,GaoLWCZ21}. 
The first one is \textit{fidelity} which evaluates how similar the surrogate models and the target models are. 
Specifically, it is defined as the percentage of the $v_i$ in $V$ where $f_{{\theta}^{'}}(v_i)=f_{\theta}(v_i)$. 
It is calculated by dividing the number of common predictions between two models by the number of the total testing inputs. 
For higher fidelity, the extracted models are expected to have more similar performance as the target models. 
Extracted models with high fidelity can be used when the attacker requires further analysis about the target models, e.g., being used in adversarial attacks as a target with infinite queries. 
Another metric is the \textit{accuracy} that represents how accurate the surrogate models are in testing data. 
Specifically, it is the percentage of the $v_i$ in $V$ where $f_{{\theta}^{'}}(v_i)=y_{v_i}$. 
It is calculated by dividing the number of the correct classified nodes by the number of the total testing nodes. 
Extracting models with higher accuracy allows the attacker to directly use them for the inference in the target application tasks rather than querying the target models, injuring the interests of model owners. 


\noindent \textbf{Models. }
Our experiments consider the case where the target model is a 2-layer graph convolution network, introduced in Eq.~\ref{eqt:GCN}.
The number of features in the hidden layer is 16.
The activation function for the hidden layer is ReLU and for the output layer is softmax.
We also apply a dropout layer with a 0.5 dropout rate after the hidden layer.
We use the Adam optimiser with a learning rate of 0.02 and training epochs of 200.
The loss function of our model is negative log-likelihood loss.

\subsection{Attack Performance}

\noindent \textbf{Overview. }
Table~\ref{tab:attack_overall} shows an overview of the performance of our seven attacks.
For Attack-0, Attack-1 and Attack-2, the numbers of the attack nodes obtained by the attacker are chosen to be about $25\%$ of the total nodes in the target networks.
For Attack-3, Attack-4, Attack-5, and Attack-6, the size of the shadow graph is set to be almost the same as the target, and the attacker is assumed to obtain fewer nodes which are about $10\%$. 
It can be found that our attacks achieve nearly equal accuracy as the target model as the baseline accuracy.
Meanwhile, most of our attacks gain about $80\%$ fidelity, which means that our extracted models mostly predict the inputs as the targets.
We highlight the attacks with the best performance among others.
Detailed discussions for each attack are presented as follow.

\noindent \textbf{Attack-0. }
Attack-0 is shown to achieve the highest fidelity since their training data is the most similar to the target model. 
We analyse their performance by adjusting several factors in the design.

\begin{figure*}
    \centering
    \begin{minipage}[htp]{0.3\linewidth}
        \centering
        \includegraphics[width=0.99\textwidth, height=0.67\textwidth]{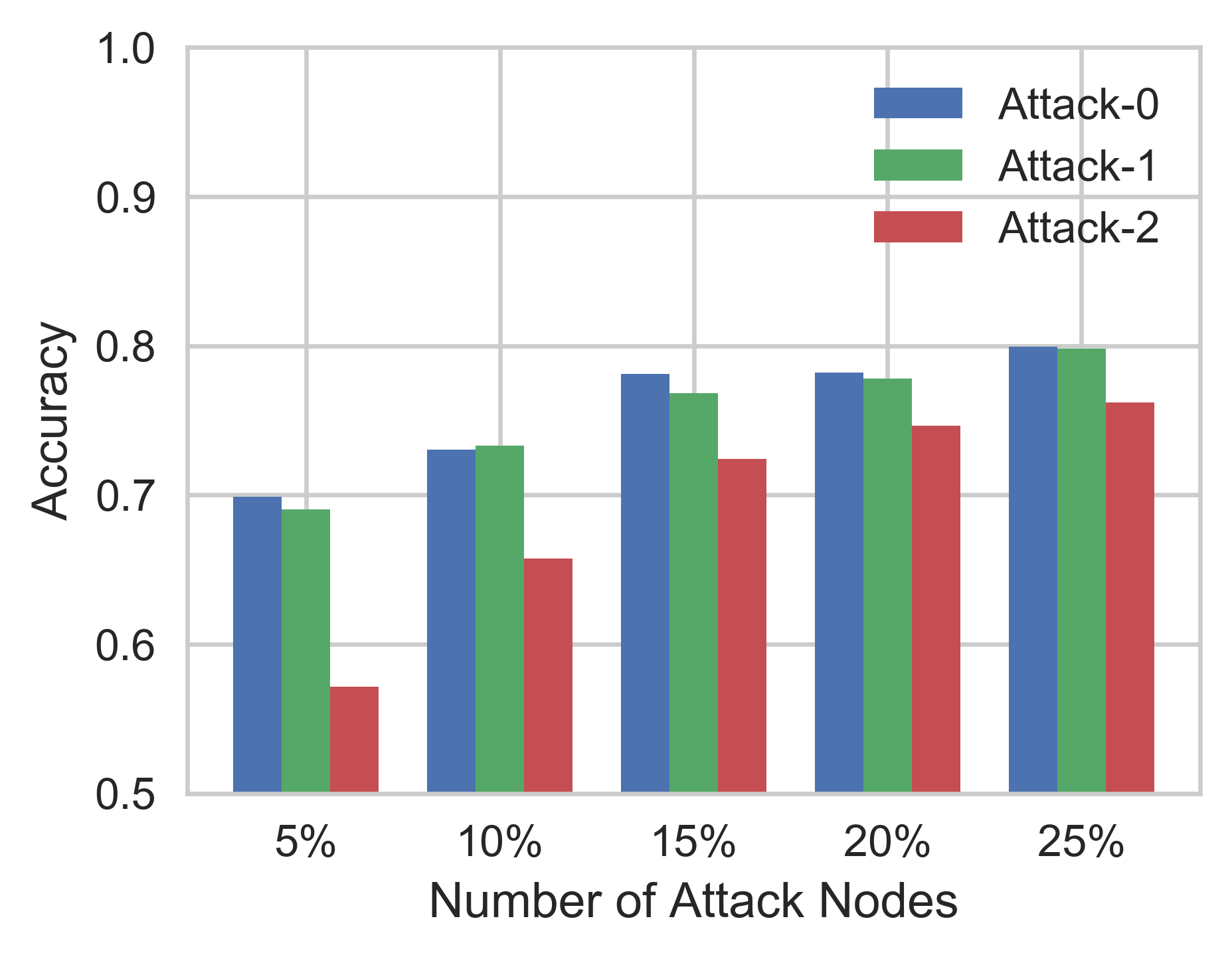} \\
        
        \includegraphics[width=0.99\textwidth, height=0.67\textwidth]{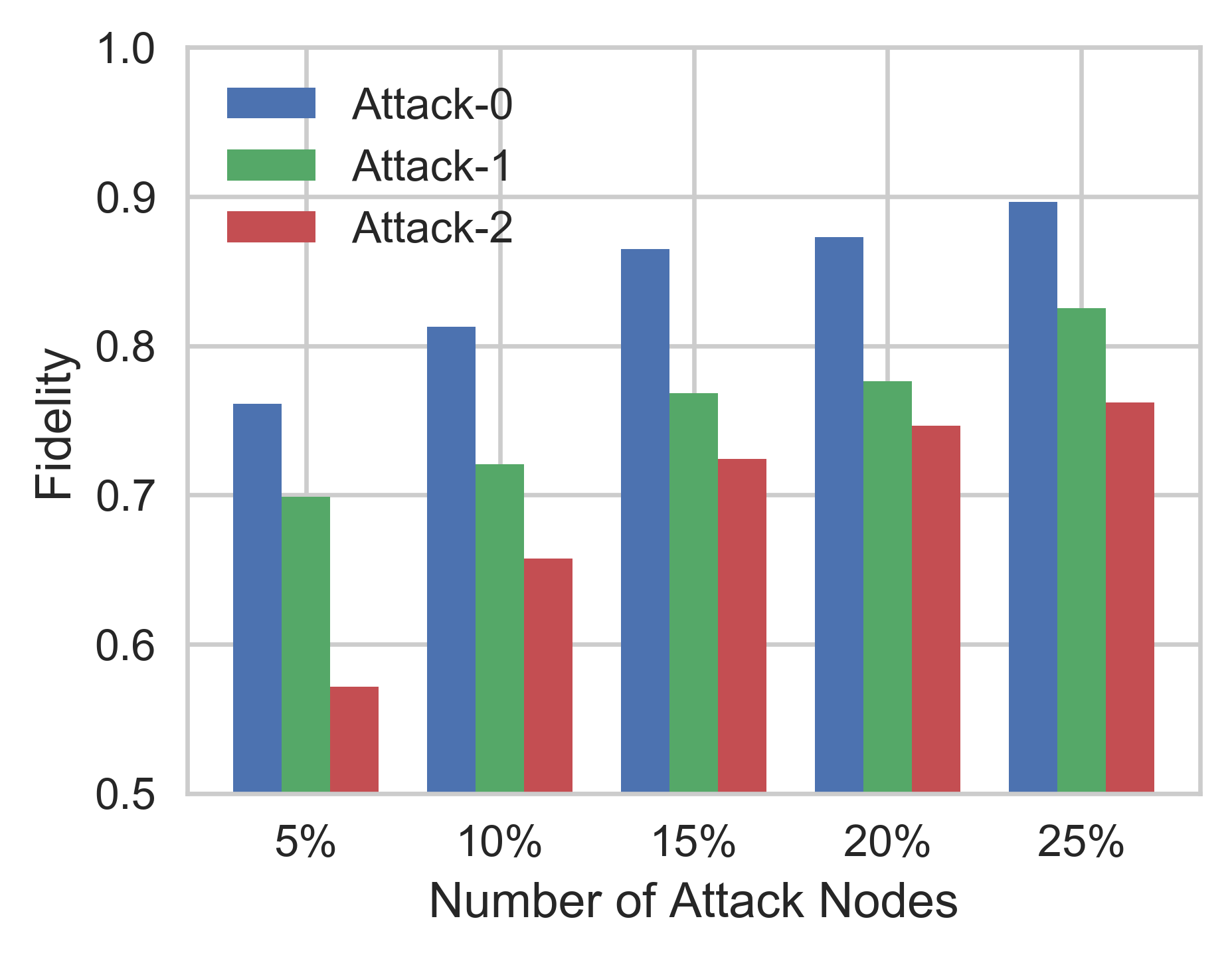}
        
        (a) Cora

    \end{minipage}
    \begin{minipage}[htp]{0.3\linewidth}
        \centering
        \includegraphics[width=0.99\textwidth, height=0.67\textwidth]{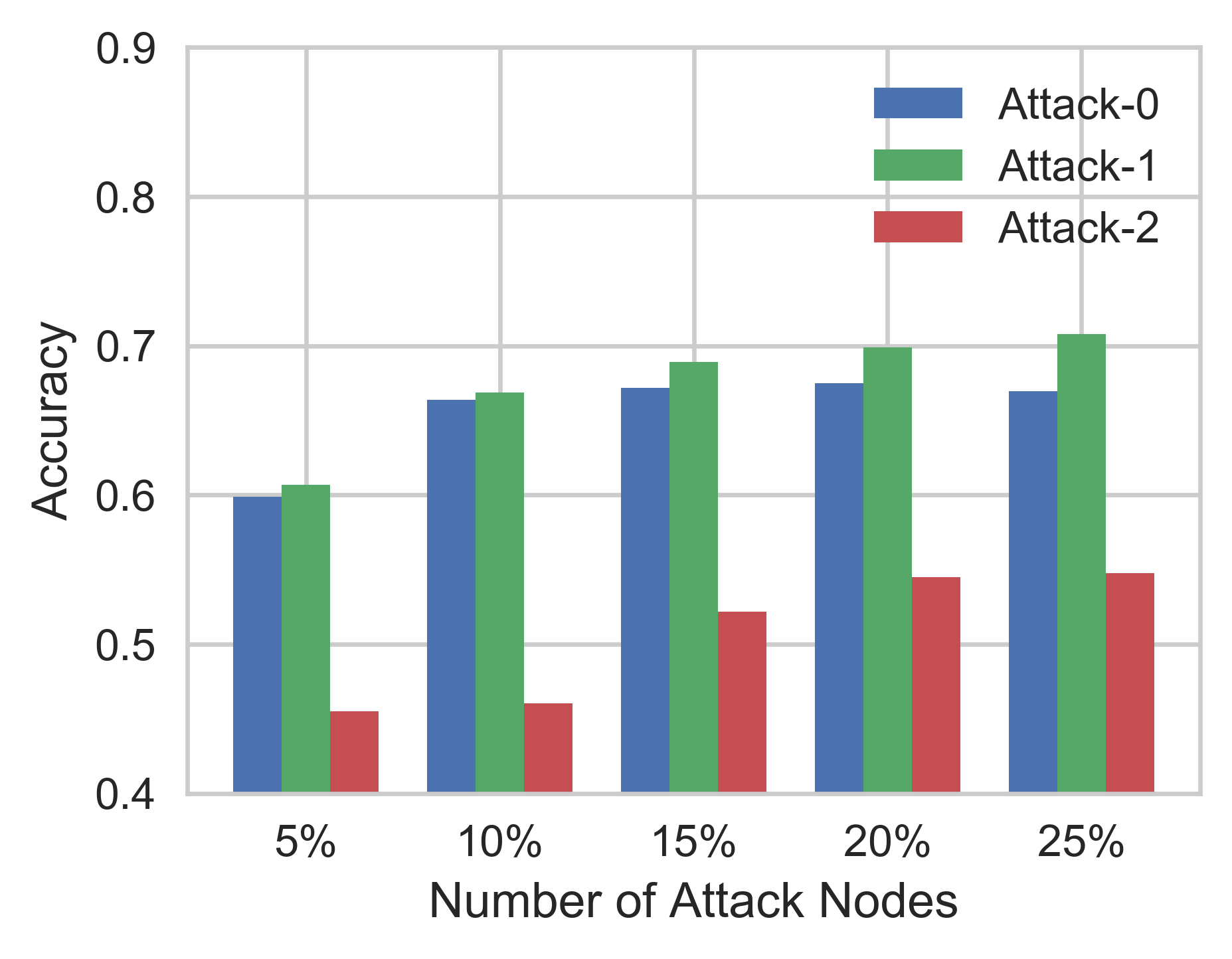}\\
        
        \includegraphics[width=0.99\textwidth, height=0.67\textwidth]{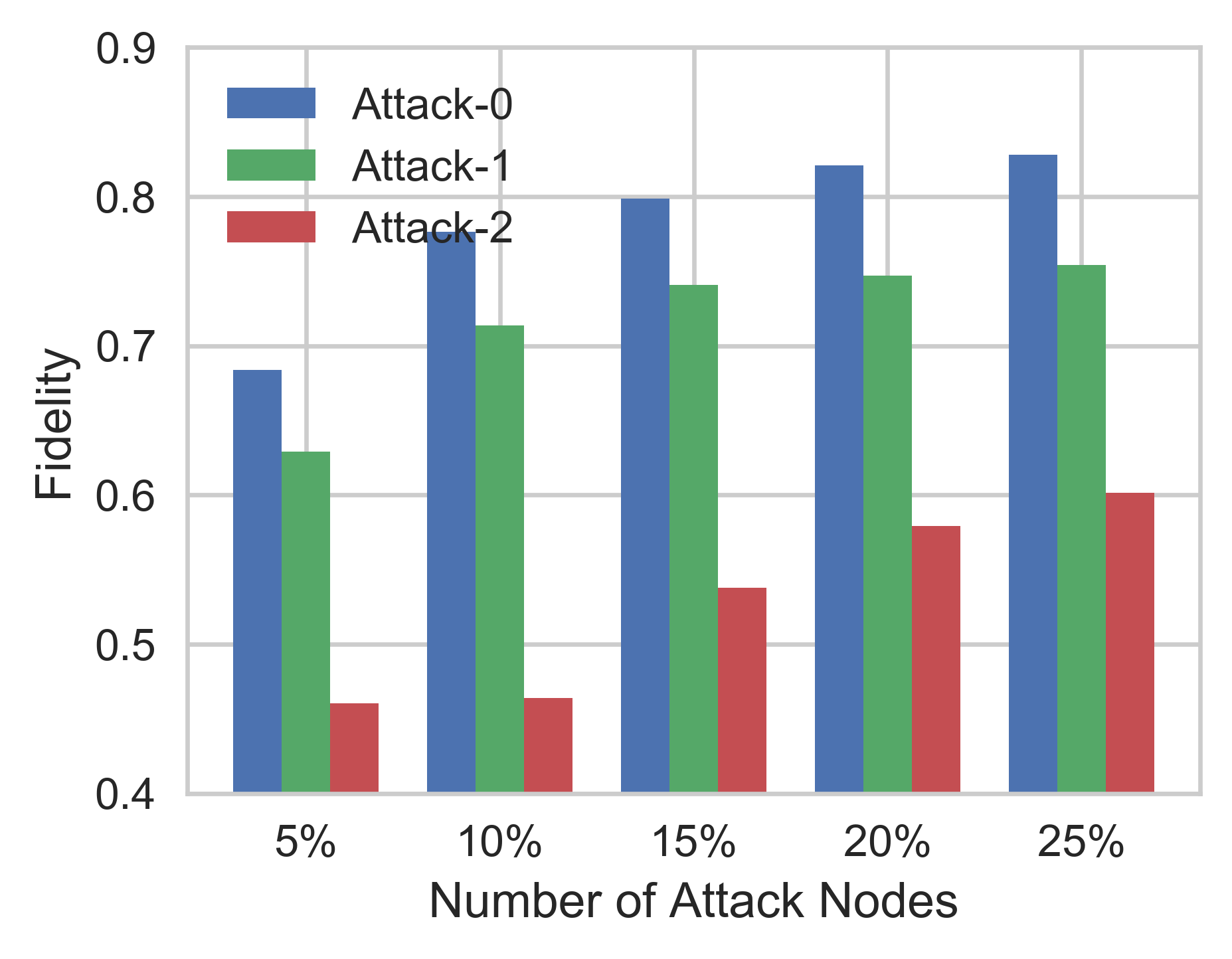}
        
        (b) Citeseer

    \end{minipage}
    \begin{minipage}[htp]{0.3\linewidth}
        \centering
        \includegraphics[width=0.99\textwidth, height=0.67\textwidth]{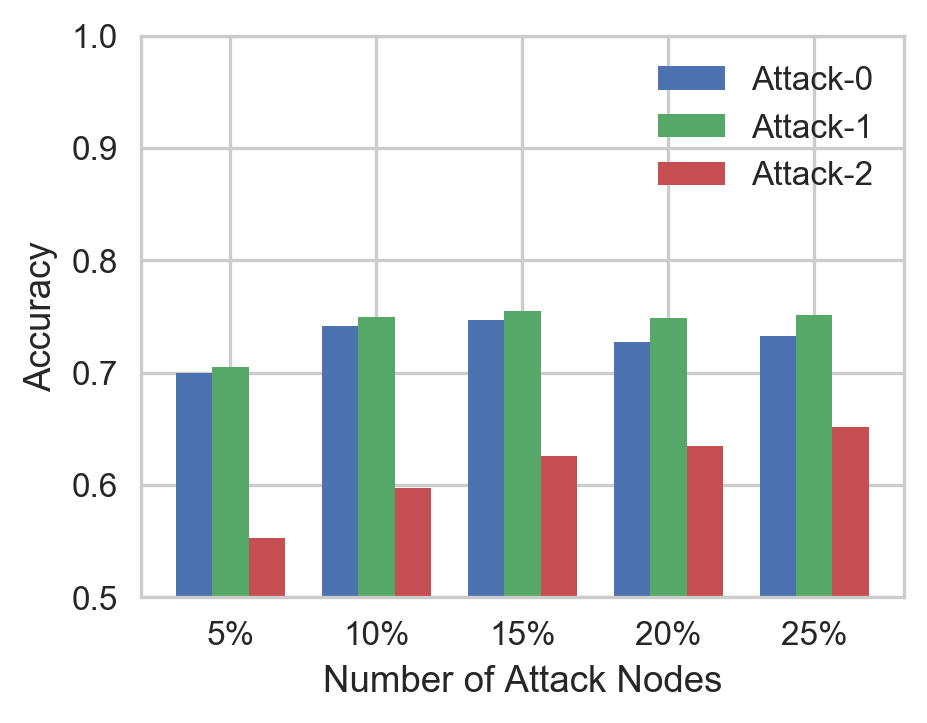} \\
        
        \includegraphics[width=0.99\textwidth, height=0.67\textwidth]{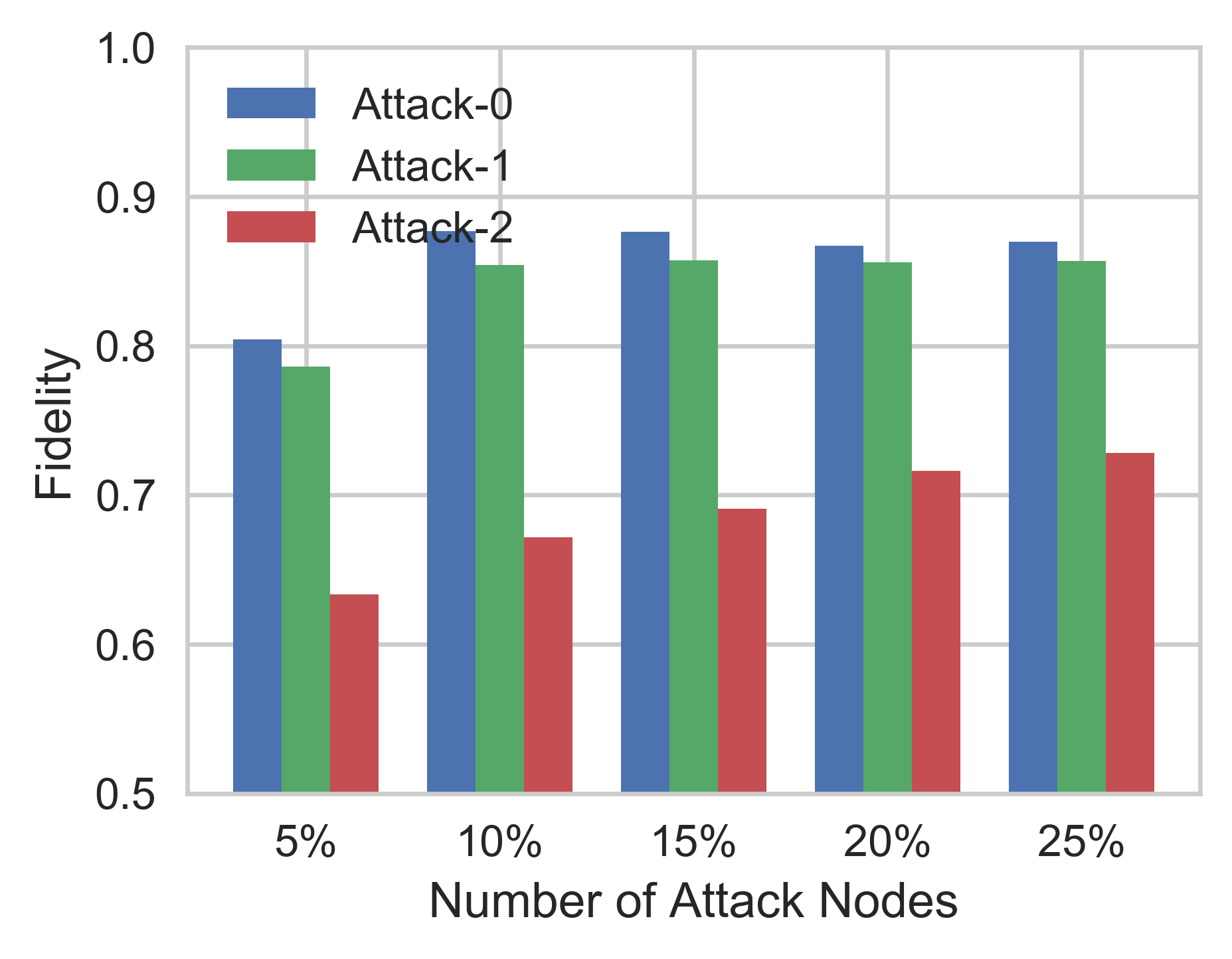}
        
        (c) Pubmed

    \end{minipage}
    \caption{Impact of the number of the attack nodes in Attack-0, Attack-1, and Attack-2}
    \label{fig:attack123_dis0}
\end{figure*}

Figure~\ref{fig:attack123_dis0} shows the relationship among the number of the attack nodes and the fidelity/accuracy of the surrogate models from $5\%$ of the total nodes to $25\%$. 
For larger numbers of attack nodes, both accuracy and fidelity increase.
The accuracy of the extracted model achieves about $79.9\%$ in the Cora dataset which is very close to the target model $81.5\%$.
And the fidelity of the duplicated model is about $90\%$.
For the Citeseer, the accuracy increases from $59.9\%$ to $67.0\%$ when obtaining the attack nodes from $5\%$ to $25\%$.
The accuracy with about $25\%$ nodes is close to the baseline accuracy $70.0\%$.
The fidelity reaches $82.8\%$ when the number of the attack node is about $25\%$ of the nodes in the total graph.
When attacking the model trained in Pubmed, the accuracy of the duplicated model increases from $70\%$ to $73\%$.

\begin{figure}[t]
    \centering
    \begin{minipage}[htp]{0.3\linewidth}
        \centering
        \includegraphics[width=0.99\textwidth, height=0.67\textwidth]{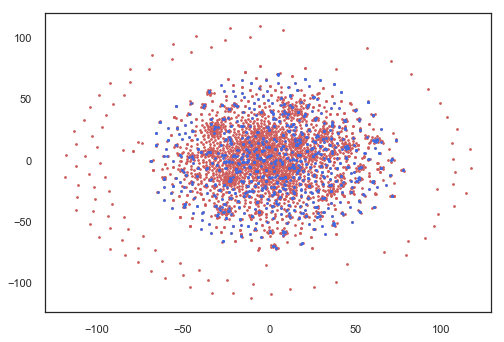} \\
        
        \includegraphics[width=0.99\textwidth, height=0.67\textwidth]{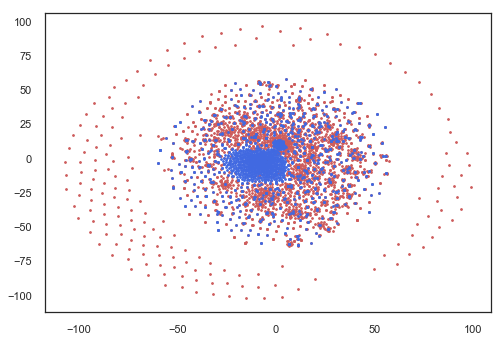}
        
        (a) Cora

    \end{minipage}
    \begin{minipage}[htp]{0.3\linewidth}
        \centering
        \includegraphics[width=0.99\textwidth, height=0.67\textwidth]{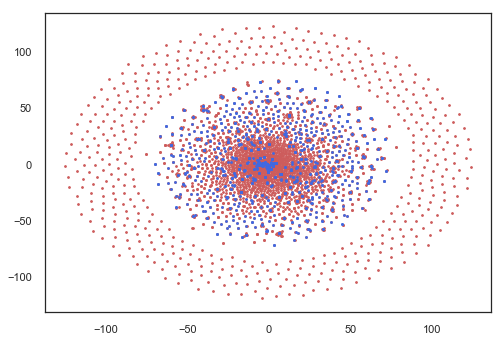} \\
        
        \includegraphics[width=0.99\textwidth, height=0.67\textwidth]{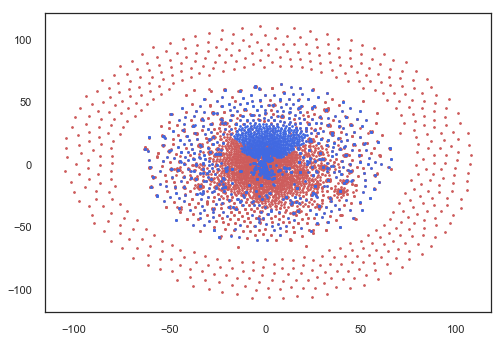}
        
        (b) Citeseer

    \end{minipage}
    \begin{minipage}[htp]{0.3\linewidth}
        \centering
        \includegraphics[width=0.99\textwidth, height=0.67\textwidth]{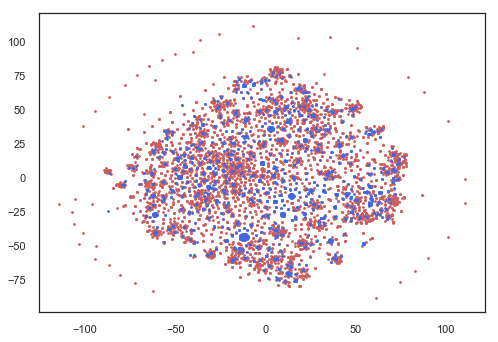} \\
        
        \includegraphics[width=0.99\textwidth, height=0.67\textwidth]{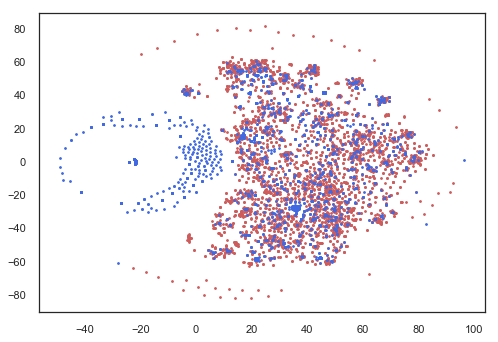}
        
        (c) Pubmed

    \end{minipage}
    \caption{Feature distribution of the nodes including only first-order synthetic neighbour nodes (Upper) and both first and second-order (Lower) for Attack-0 projected into a 2-dimension space using t-SNE. }
    \label{fig:attack0_dis0}
\end{figure}

We also evaluate how synthesising the neighbours affects our attack performance.
Table~\ref{tab:attack_0_dis_1} shows the accuracy and fidelity of the attack with and without the synthetic nodes.
It can be found that, synthesising the attributes for the neighbours of the attack nodes can improve the fidelity of our attacks.
We also evaluate the attack performance when synthesising more neighbour nodes.
It is shown that too many synthetic nodes will hurt our attacks.
We compare the feature distribution of the graph generating by different strategies in Figure~\ref{fig:attack0_dis0}. 
It is shown that the graph generated by synthesising only the first order achieves the most similar distribution as the target graph that matches our attack results. 
We also compare the degree distribution in Appendix~\ref{app:attack0_dis0}. 

\begin{figure}[t]
    \small
    \centering
    \includegraphics[width=0.49\textwidth, height=0.23\textwidth]{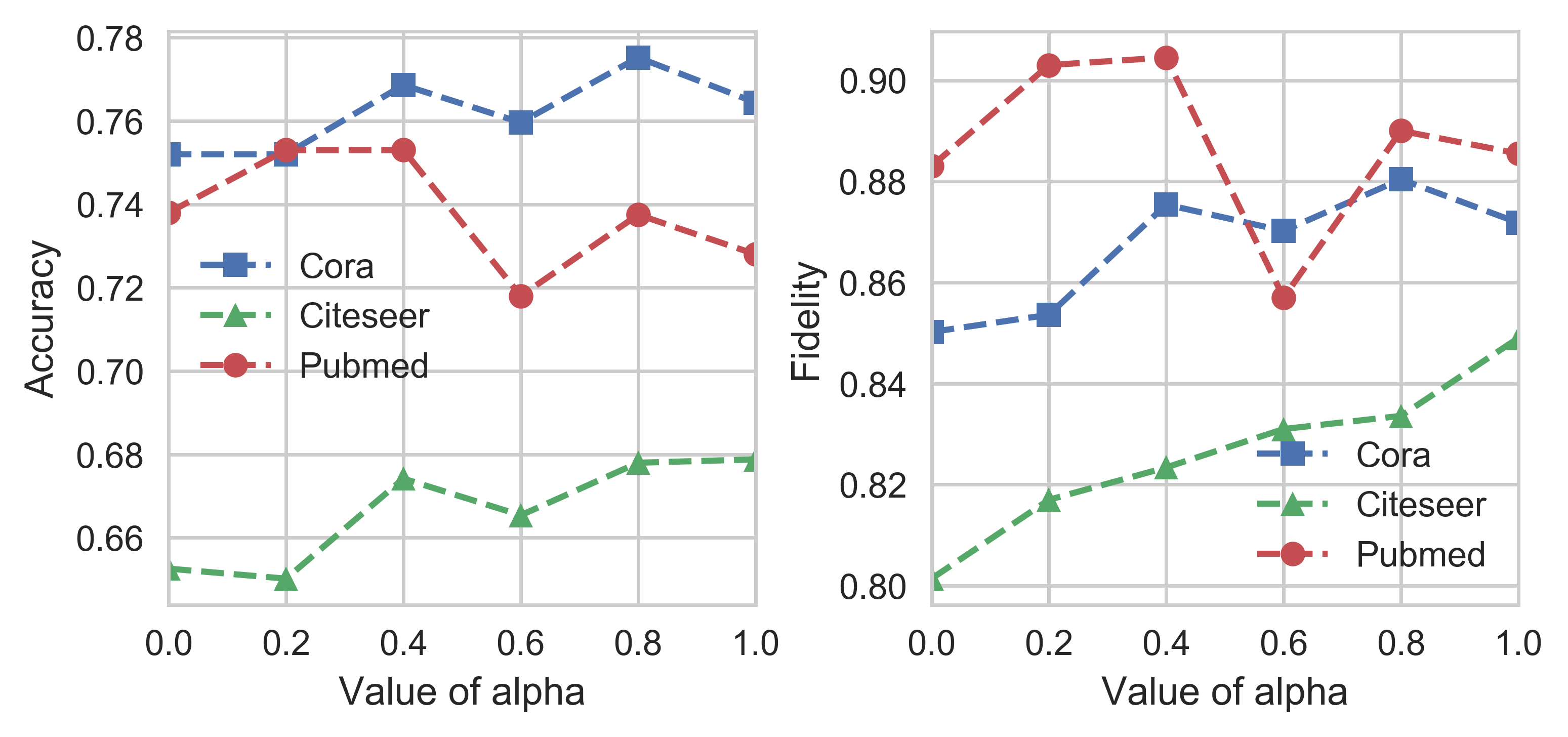}
    \caption{Impact of the adjustment factor $\alpha$ for Attack-0}
    \label{fig:attack_0_impact_alpha}
\end{figure}

We now discuss the impact of the adjustment factor $\alpha$.
Figure~\ref{fig:attack_0_impact_alpha} shows both the accuracy and fidelity of the attack with variant $\alpha$.
The experiments show that this factor does affect the attack performance but mostly inside $\pm 5\%$.
We can also find that the attack performance raises when $\alpha$ increases for both Cora and Citeseer.
Larger $\alpha$ means the synthetic attributes of the nodes are more based on their 1-hop neighbours. 
This is reasonable since the relationship between the synthetic nodes to their 1-hop neighbours is stronger than the 2-hop neighbours. 
Meanwhile, the performance from Pubmed is undulate.
To achieve the best attack performance, the attacker can carefully choose the adjustment factor by considering the characteristic of the graph.

\begin{table*}[t]
\small
\centering
\begin{tabular}{c|c|c|c|c}
    \hline
        Metric & Dataset & Without Synthetic & First Order Neighbour Synthetic  & Second Order Neighbour Synthetic \\
    \hline
        \multirow{3}{*}{Accuracy}    & Cora & $0.797\pm0.012$ & \bm{$0.799\pm0.009$} & $0.793\pm0.008$ \\ 
    \cline{2-5} 
          & Citeseer & \bm{$0.688\pm0.013$} & $0.684\pm0.016$ & $0.681\pm0.017$ \\  
    \cline{2-5} 
          & Pubmed & $0.735\pm0.027$ & \bm{$0.736\pm0.004$} & $0.731\pm0.013$ \\  
    \hline
        \multirow{3}{*}{Fidelity} & Cora & $0.869\pm0.012$ & \bm{$0.896\pm0.008$} & $0.889\pm0.009$ \\   
    \cline{2-5} 
          & Citeseer & $0.816\pm0.030$ & \bm{$0.848\pm0.019$} & $0.834\pm0.015$ \\  
    \cline{2-5} 
          & Pubmed & $0.879\pm0.030$ & \bm{$0.890\pm0.007$} & $0.886\pm0.015$ \\  
    \hline
\end{tabular}
\caption{Impact of the synthetic nodes for Attack-0. Best results are highlighted in bold. }
\label{tab:attack_0_dis_1}
\end{table*}

\noindent \textbf{Attack-1. }
In Attack-1, only the attributes of the attack nodes are known to the attackers while their connections are unknown.
Therefore, we generate the graph structure based on these node features.
Figure~\ref{fig:attack123_dis0} shows the relationship between the number of attack nodes and the attack performance.
Similar to Attack-0, more attack nodes can significantly increase the accuracy and fidelity of the extracted models.

To clearly show how the graph structure generation contributes to our design, we set a baseline, which uses deep neural networks to infer the output labels based only on the input features.
The results for both accuracy and fidelity of the surrogate models in three different datasets are shown in Table~\ref{tab:attack2_dis0}.
In the Cora dataset, our design achieves about $79.8$ for the accuracy of the extracted model while DNNs have only $57.7\%$.
And our attack improves the fidelity from $59.0\%$ to $82.5\%$. 

To show how the generated graph structure matches the original graph, we also evaluate the degree distribution of the graph generated by our attack methods.
Notice that the attribute distribution should be very similar to the target graph since the attributes in Attack-1 are all from the attack nodes which is considered as a sample of the original.
The comparisons between the degree distribution of the generated graph and the target graph for three datasets are shown in the Appendix~\ref{app:attack1_dis0}. 
We show that they are more similar comparing with the distribution without graph generation method. 
This demonstrates that using graph structure generation is a good approach to help reconstruct the graph structure and further improve our attack.


\noindent \textbf{Attack-2. }
This attack considers the scenario when the attacker has only knowledge about the graph structure.
The results for both accuracy and fidelity are shown in Table~\ref{tab:attack_overall}. 
And Figure~\ref{fig:attack123_dis0} shows how the number of the attack nodes affects our attack.
Similarly, both accuracy and fidelity increase when obtaining more attack nodes.

Notice that, Attack-2 has the worst performance comparing with Attack-0 and Attack-1. 
This might cause by the less similarity of the synthetic items for the attack graph we generated. 
For other attacks, our design can generate node attributes or connections similar to the target graph.
However, due to the lack of knowledge about the attributes, the synthetic attributes of the attack graph for this type of attack are one-hot vectors that are far from the target. 
If the attacker can obtain some knowledge about the node attributes which can be used to synthesise similar attributes, the performance can be improved.


\noindent \textbf{Attack-3. }
For Attack-3, we consider the case where the attacker can only have a shadow graph defined in Section~\ref{sec:knowledge} without knowing both attributes and graph structures for the target models.
The results for both accuracy and fidelity are shown in Table~\ref{tab:attack_overall}.
Compared to previous attacks with some background knowledge of the target graph, the accuracy of this type of attack is similar or even better. 
Thus, obtaining the complete graph data can achieve high accuracy if the shadow graph has the same domain as the target.
However, the fidelity of the attack is significantly smaller than previous attacks since the training graph data of the target model is entirely different from our extracted graph.
Our target model is built as the GCN model which is transductive, so it is hard to gain an extracted model with similar functionality. 

\begin{figure}
    \centering
    \includegraphics[width=0.49\textwidth, height=0.23\textwidth]{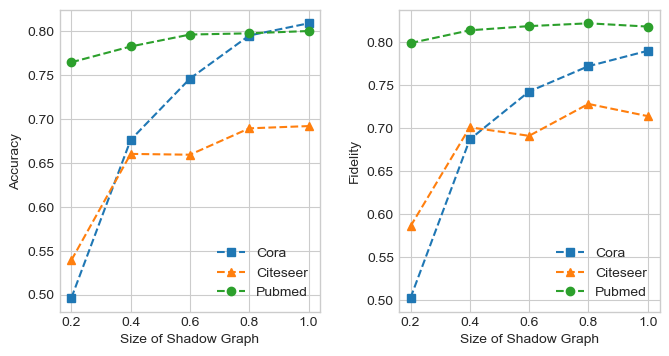}
    \caption{Impact of the shadow graph size for Attack-3}
    \label{fig:attack_3_dis0}
\end{figure}

We also analyse the effect for different knowledge of the shadow sub-graph.
Figure~\ref{fig:attack_3_dis0} shows the relationship between the attack performance and the size of shadow graph.
The x-axis represents the ratio of the size of the shadow graph to the target graph.
It can be found that while knowing the larger size of the shadow graph, the accuracy of the surrogate models increases a lot.
It is obvious since the attacker can extract more knowledge from a larger training graph.
It can also be found that the fidelity of the surrogate models becomes saturated even the shadow graph size becomes larger. 
As discussed, the target GCN model is transductive, which makes the attacker difficult to obtain a highly equivalent model.
Therefore, the fidelity of our attack will reach the ceiling when the size of the shadow graph keeps increasing.


\begin{table}[t]
\centering
\begin{tabular}{c|c|c|c}
    \hline
        Metric & Dataset &  Simple DNN & LDS-GNN~\cite{Franceschi19} \\
    \hline
        \multirow{3}{*}{Accuracy} & Cora & $0.577\pm0.004$ & \bm{$0.798\pm0.006$}\\ 
    \cline{2-4} 
          & Citeseer & $0.596\pm0.004$ & \bm{$0.708\pm0.007$} \\  
    \cline{2-4} 
          & Pubmed & $0.727\pm0.006$ & \bm{$0.751\pm0.003$} \\  
    \hline
        \multirow{3}{*}{Fidelity} & Cora & $0.590\pm0.010$ & \bm{$0.825\pm0.007$}\\   
    \cline{2-4} 
          & Citeseer & $0.632\pm0.005$ & \bm{$0.754\pm0.005$} \\  
    \cline{2-4} 
          & Pubmed & $0.761\pm0.005$ & \bm{$0.857\pm0.003$} \\  
    \hline
\end{tabular}
\caption{Fidelity/accuracy for Attack-1. Best results are highlighted in bold. }
\label{tab:attack2_dis0}
\end{table}

\begin{figure*}
    \centering
    \begin{minipage}[htp]{0.3\linewidth}
        \centering
        \includegraphics[width=0.99\textwidth, height=0.67\textwidth]{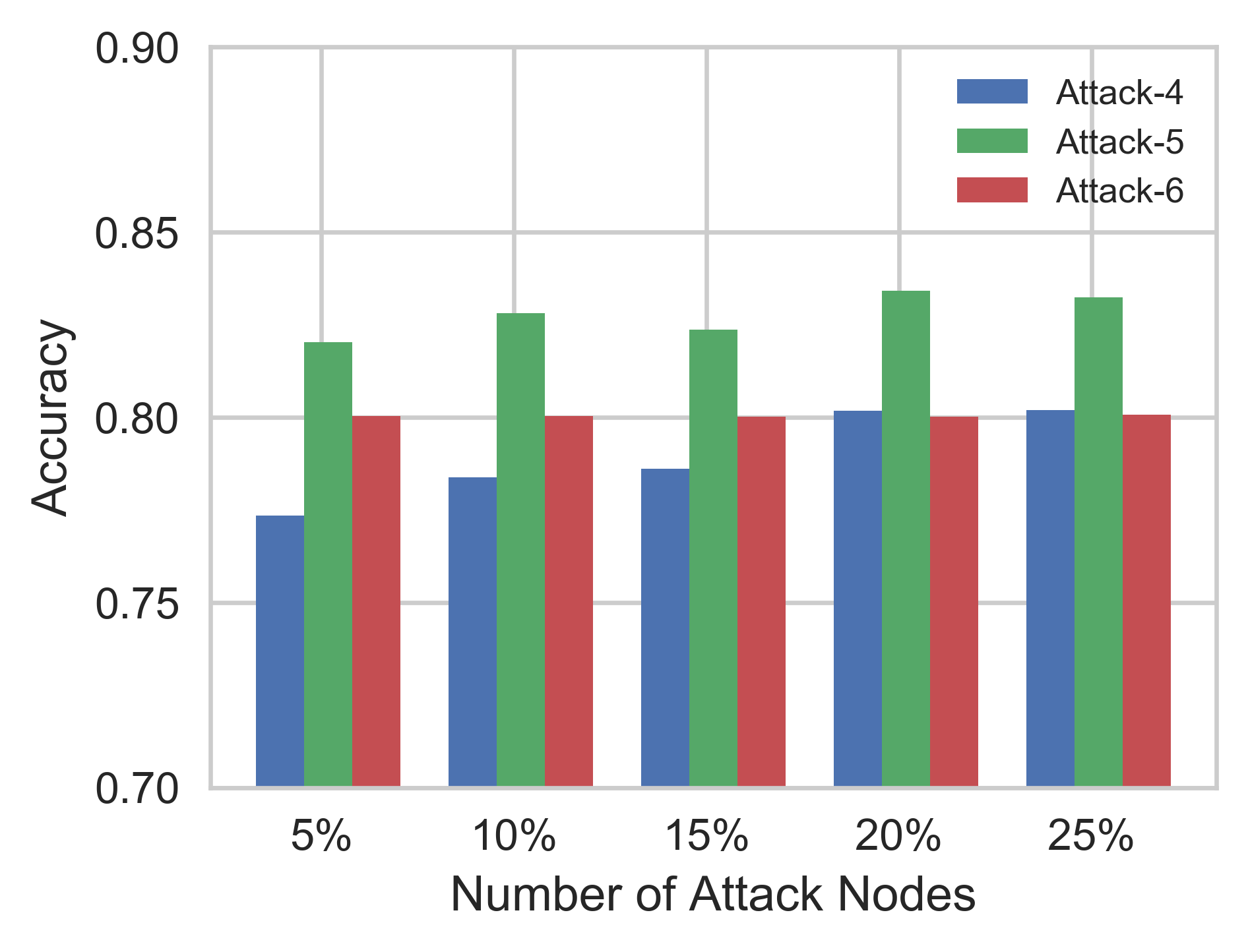} \\
        
        \includegraphics[width=0.99\textwidth, height=0.67\textwidth]{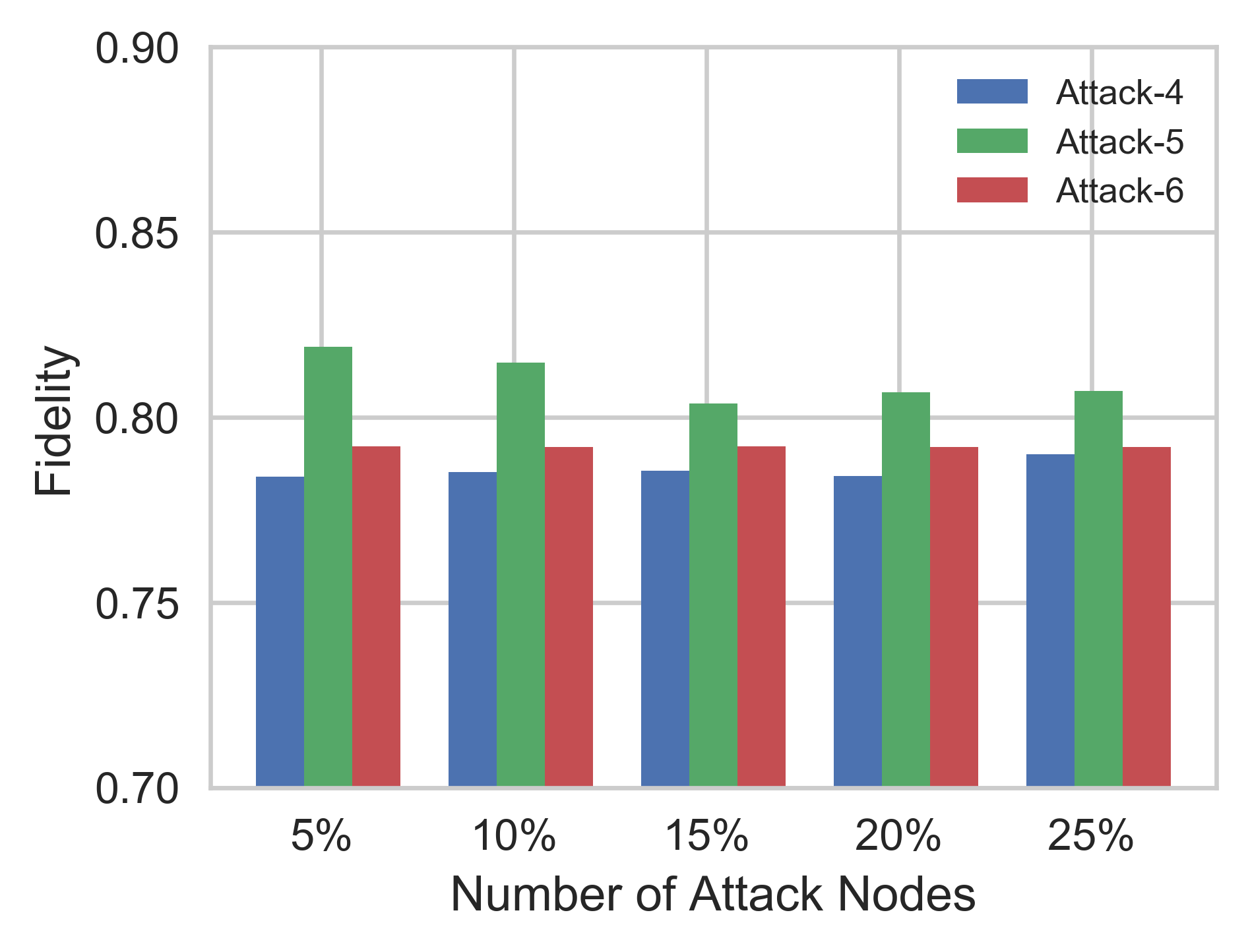}
        
        (a) Cora

    \end{minipage}
    \begin{minipage}[htp]{0.3\linewidth}
        \centering
        \includegraphics[width=0.99\textwidth, height=0.67\textwidth]{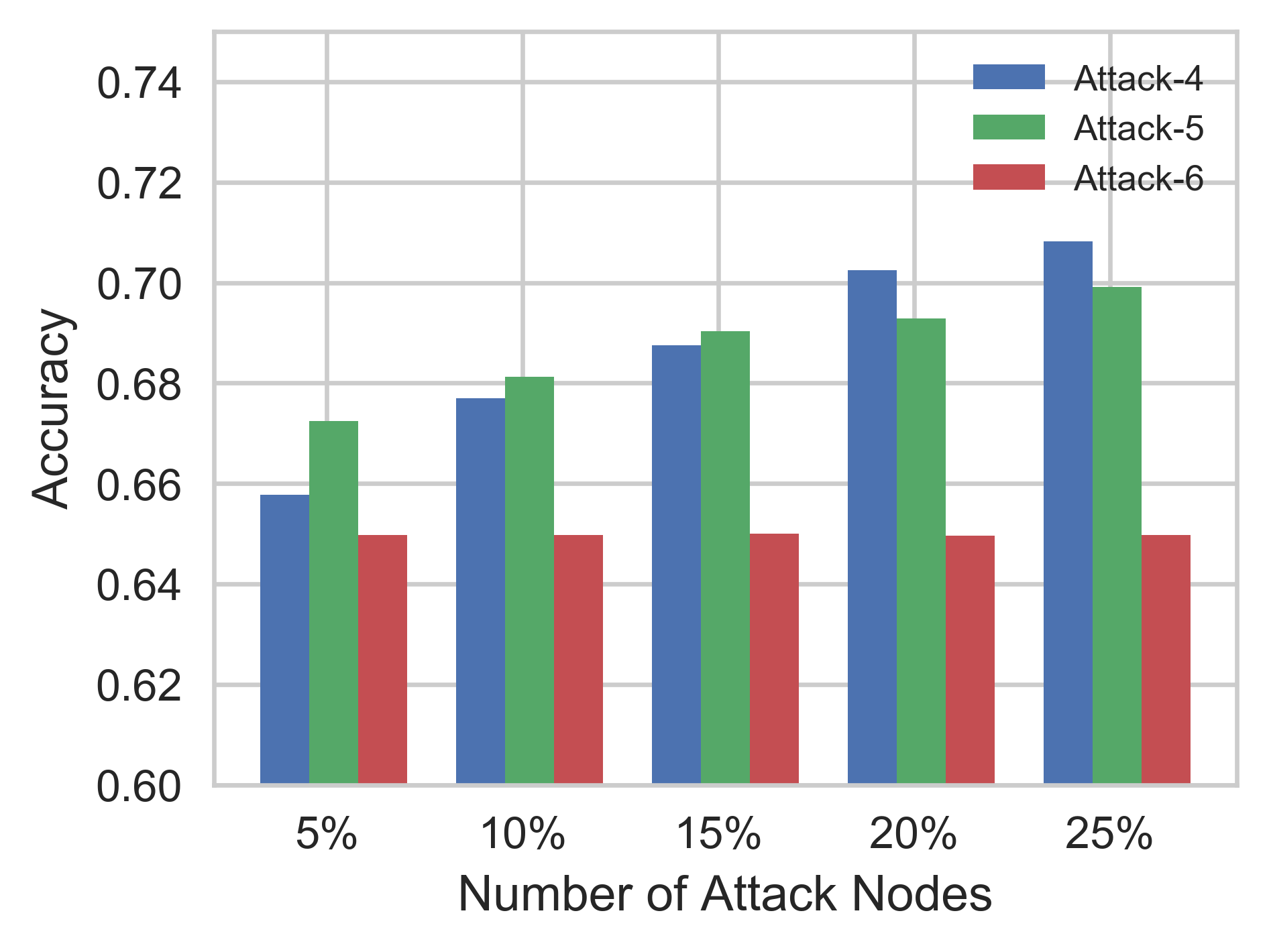}\\
        
        \includegraphics[width=0.99\textwidth, height=0.67\textwidth]{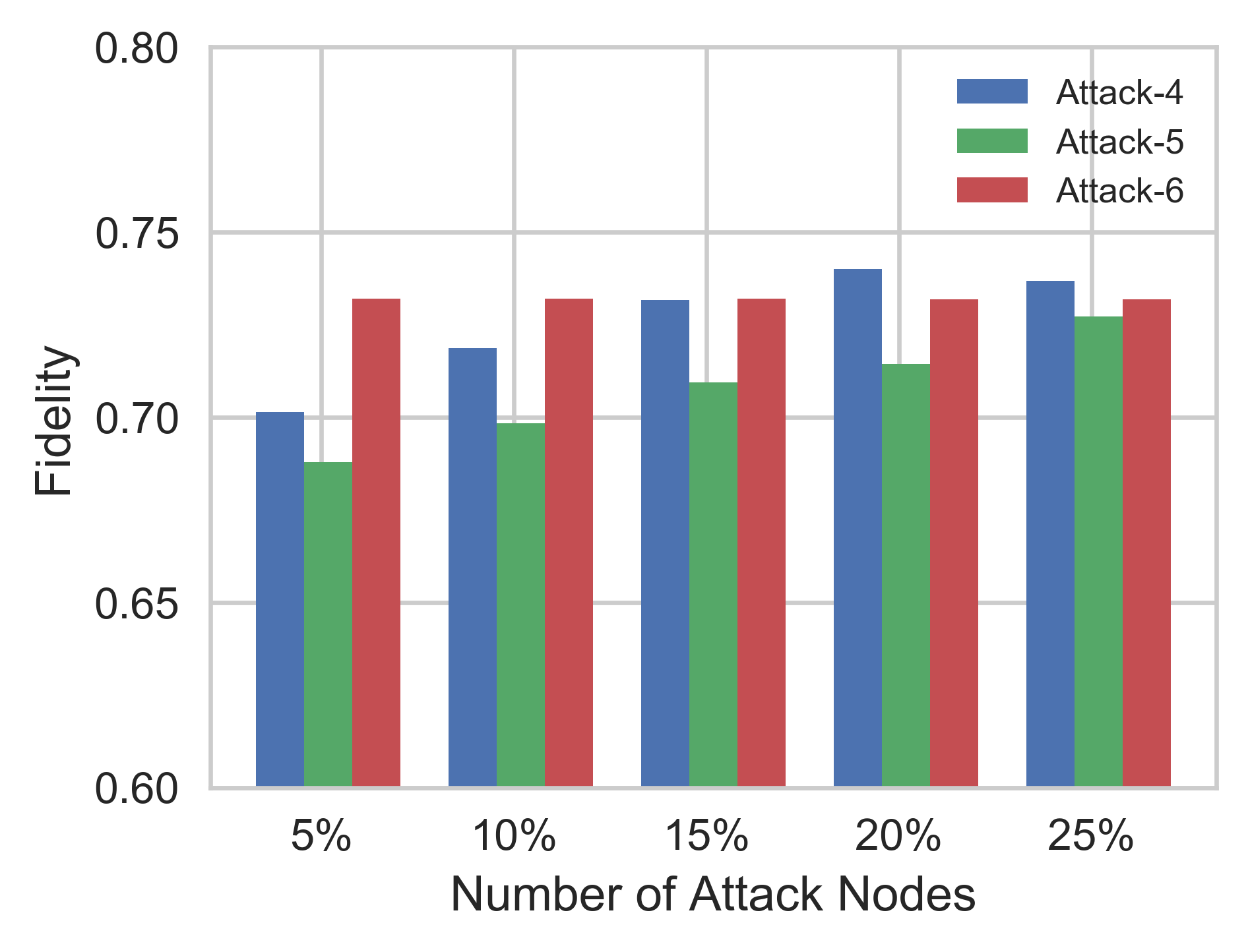}
        
        (b) Citeseer

    \end{minipage}
    \begin{minipage}[htp]{0.3\linewidth}
        \centering
        \includegraphics[width=0.99\textwidth, height=0.67\textwidth]{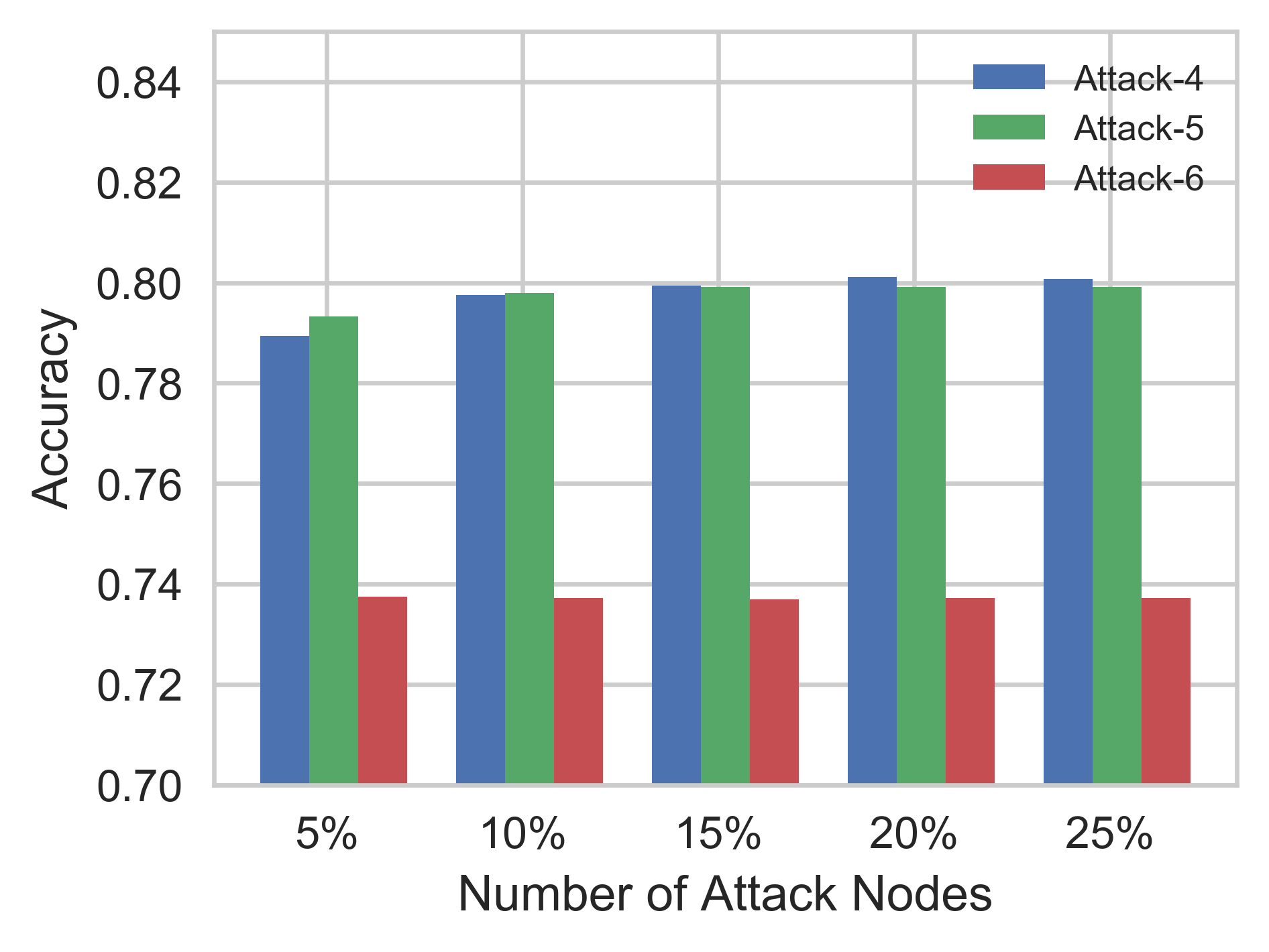} \\
        
        \includegraphics[width=0.99\textwidth, height=0.67\textwidth]{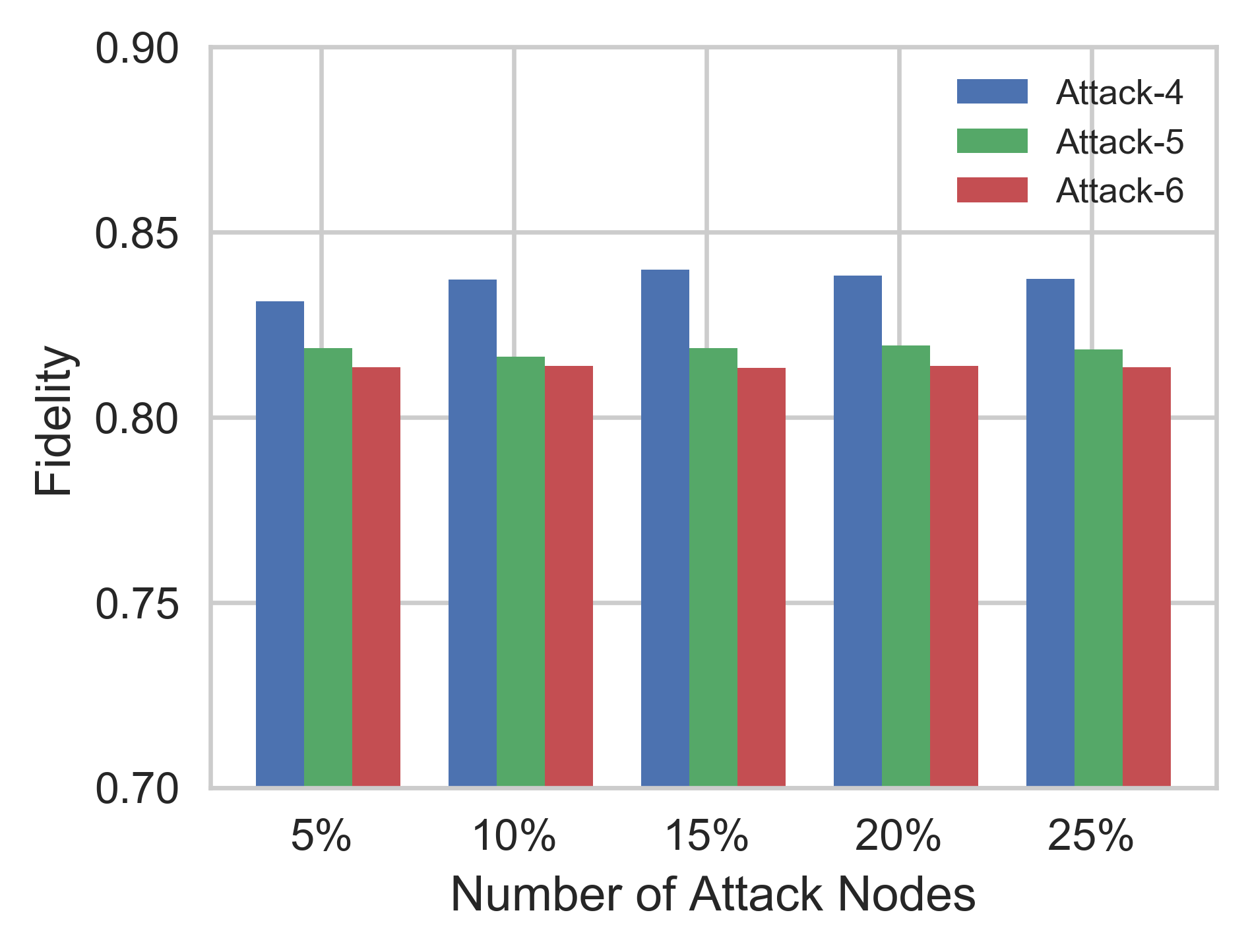}
        
        (c) Pubmed

    \end{minipage}
    \caption{Impact of the number of the attack nodes in Attack-4, Attack-5, and Attack-6}
    \label{fig:attack456_dis0}
\end{figure*}

\noindent \textbf{Attack-4. }
Now we consider the case when the attacker has access to a shadow graph as well as some attack nodes in the target graph.
Based on Table~\ref{tab:attack_overall}, it can be found that Attack-4 achieves higher accuracy and fidelity than Attack-3. 
It demonstrates that obtaining extra knowledge can lead to better attack performance.
Note that, the improvement of the fidelity is more significant than the accuracy especially for the Citeseer and Pubmed dataset.
This shows that the knowledge from the target graph can be helpful to extract models with similar predictions as the target model.

We also evaluate how the number of the attack nodes affects the attack performance. 
Figure~\ref{fig:attack456_dis0} shows the accuracy and fidelity for different numbers of attack nodes. 
Compared with Attack-0 which does not know the shadow graph, both accuracy and fidelity are only slightly affected by the numbers of the attack nodes. 
Note that, the number of the attack nodes and their 2-hop neighbours are commonly smaller than the number of nodes in shadow graph. 
Thus, learning from the combination of both of them, the performance of Attack-4 is hardly affected by the number of the attack nodes as the former attacks (Attack-0/1/2). 




\noindent \textbf{Attack-5. }
Now we consider the case when the attacker has access to a shadow graph and also some attack nodes in the target graph.
Based on Table~\ref{tab:attack_overall}, it can be found that this attack achieves slightly higher accuracy and fidelity than the Attack-3 and nearly equal attack performance as the Attack-4.
But since the knowledge of Attack-5 is less than Attack-4 (the neighbour graph structure is unknown), the fidelity is lower which is similar to the comparison between Attack-0 and 1.
It demonstrates that obtaining more background knowledge can enhance our extraction attacks more.
%
Similar to Attack-4, the number of the attack nodes can only slightly increases the accuracy and fidelity of the attacks, as shown in Figure~\ref{fig:attack456_dis0}. 
\noindent \textbf{Attack-6. }
Finally, we discuss the attack when the attacker has both knowledges about the graph structure and the shadow graph.
With the help of shadow graph, the overall performance of Attack-6 is significantly higher than Attack-2.
It demonstrates that introducing knowledge about the node attributes can improve the attack performance.
It also achieves comparative performance as the Attack-4 and Attack-5.

When assessing the impact from the attack nodes, we find that increasing the number of the attack nodes can not affect the attack performance, as shown in Figure~\ref{fig:attack456_dis0}. 
This is caused by the different attack performance of the two component models in the ensemble models for Attack-6. 
Based on our design, the ensemble models infer the node label based on the posteriors of two extracted models.
Since the component model built as Attack-3 achieves both higher accuracy and fidelity than the model for Attack-2, the ensemble models learn more from the model built as Attack-3. 
Using more attack nodes and increasing the accuracy of the model for Attack-2 can hardly impact the overall attack performance. 
\section{Conclusion and Future Work}
\label{sec:conclusion}
In this paper, we demonstrate a model extraction attack against GNNs.
We first generate legitimate-looking queries as the normal nodes among the target graph, then utilise the query responses and accessible structure knowledge to reconstruct the model.
We characterise the problem into seven threat models considering different knowledge of the attacker.
Then we accordingly propose seven attacks based on the knowledge and the query responses.
Our experimental results show that our attack obtains surrogate models with similar predictions as the targets.
%

An emerging research direction is a defence against the extraction attacks in GNNs. 
Existing defences against the model extraction attacks on DNN system propose to monitor and filter the input queries~\cite{Juuti19,Kesarwani18}, which can be extended and combined wit the structure analysis when implementing the defence for GNNs. 
We consider it as our future work. 

\section*{Acknowledgment}
This research was supported by the Australian Research Council (ARC) under a Future Fellowship No. FT210100097.

\bibliographystyle{ACM-Reference-Format}
\bibliography{GNN-ME}

\newpage
\appendix

\section{Appendix}
\subsection{Degree distribution for Attack-0. }
\label{app:attack0_dis0}
\begin{figure}[H]
    \centering
    \begin{minipage}[htp]{0.8\linewidth}
        \centering
        \includegraphics[width=0.99\textwidth, height=0.67\textwidth]{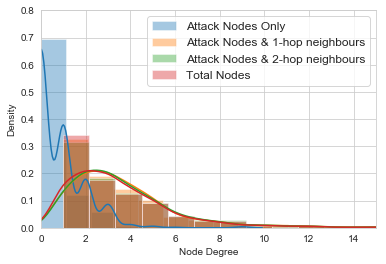}
        
        (a) Cora

    \end{minipage}
    \begin{minipage}[htp]{0.8\linewidth}
        \centering
        \includegraphics[width=0.99\textwidth, height=0.67\textwidth]{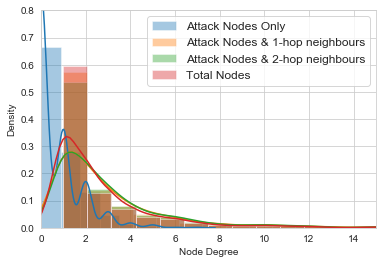}
        
        (b) Citeseer

    \end{minipage} \\
    \begin{minipage}[htp]{0.8\linewidth}
        \centering
        \includegraphics[width=0.99\textwidth, height=0.67\textwidth]{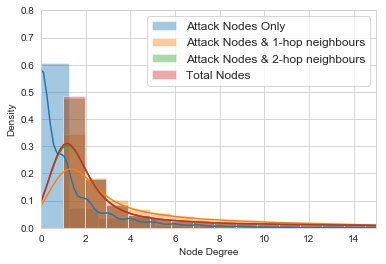}
        
        (c) Pubmed

    \end{minipage} \\
    \caption{Degree Distribution for Attack-0}
    \label{fig:attack0_dis1}
\end{figure}

We evaluate the degree distribution when using different synthesise methods for Attack-0 as shown in Figure~\ref{fig:attack0_dis1}. 
We compare the distribution for the total nodes in the target graph, the attack nodes with their 1-hop or 2-hop neighbours, and only the attack nodes. 
It can be observed that, utilising the neighbours of the attack nodes can significantly synthesise the distribution of the entire graph. 
Therefore, it is necessary to consider the neighbours of the attack nodes and synthesise their attributes if they are inaccessible to the attackers. 

\subsection{Degree distribution for Attack-1. }
\label{app:attack1_dis0}
\begin{figure}[H]
    \centering
    \begin{minipage}[htp]{0.8\linewidth}
        \centering
        \includegraphics[width=0.99\textwidth, height=0.67\textwidth]{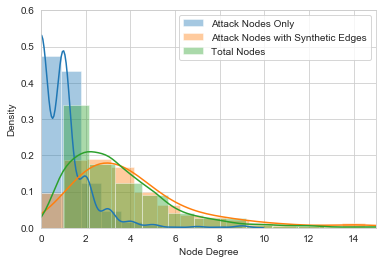}
        
        (a) Cora

    \end{minipage}
    \begin{minipage}[htp]{0.8\linewidth}
        \centering
        \includegraphics[width=0.99\textwidth, height=0.67\textwidth]{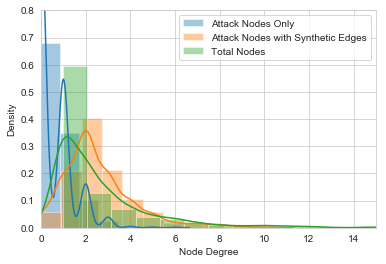}
        
        (b) Citeseer

    \end{minipage} \\
    \begin{minipage}[htp]{0.8\linewidth}
        \centering
        \includegraphics[width=0.99\textwidth, height=0.67\textwidth]{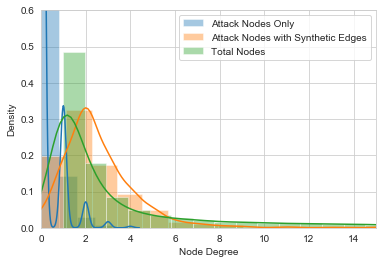}
        
        (c) Pubmed

    \end{minipage}
    \caption{Degree Distribution for Attack-1}
    \label{fig:attack1_dis0}
\end{figure}

We also evaluate the degree distribution with/without synthesising methods for Attack-1 as shown in Figure~\ref{fig:attack1_dis0}. 
We compare the distribution among the total nodes in the target graph, the attack nodes with synthetic edges, and only the attack nodes. 
It can be found that, once applying the graph structure generation methods, the synthesised graph can be more similar to the target graph and benefits our extraction attacks. 

\end{document}